\documentclass[11pt]{article}

\usepackage[preprint]{acl}

\usepackage{times}
\usepackage{latexsym}
\usepackage[T1]{fontenc}
\usepackage[utf8]{inputenc}
\usepackage{microtype}
\usepackage{inconsolata}
\usepackage{graphicx}
\usepackage{amsmath}
\usepackage{booktabs}
\usepackage{colortbl}
\usepackage{float}

\definecolor{accblue}{HTML}{2A78D6}
\definecolor{conforange}{HTML}{EB6834}
\definecolor{gemma}{HTML}{6250D6}
\definecolor{qweneight}{HTML}{EDA100}
\definecolor{qwenfive}{HTML}{1BAF7A}
\definecolor{ensemble}{HTML}{8A8983}
\newcommand{\basedot}[1]{\textcolor{#1}{\rule[0.1ex]{0.55em}{0.55em}}}
\newsavebox{\fitbox}
\newcommand{\fitwidth}[1]{\sbox{\fitbox}{#1}\ifdim\wd\fitbox>\linewidth\resizebox{\linewidth}{!}{\usebox{\fitbox}}\else\usebox{\fitbox}\fi}

\title{Lost in Translation, Not in Confidence:\\ Benchmarking and Calibrating Multilingual Decision Models}

\author{
  \textbf{Mingzhe Du\textsuperscript{1,2},}
  \textbf{Thong Thanh Nguyen\textsuperscript{2},}
  \textbf{Nguyen Tran Cong Duy\textsuperscript{1},}
  \textbf{See-Kiong Ng\textsuperscript{2},}
  \textbf{Luu Anh Tuan\textsuperscript{1,3}}
  \\[0.5ex]
  \textsuperscript{1}Nanyang Technological University \quad
  \textsuperscript{2}National University of Singapore \quad
  \textsuperscript{3}VinUniversity
  \\
  \texttt{\{mingzhe001, nguyentr003, anhtuan.luu\}@ntu.edu.sg}, \\ \texttt{thong.nguyen@nus.edu.sg}, \texttt{seekiong@nus.edu.sg}
}

\hypersetup{pdftitle={Lost in Translation, Not in Confidence: Benchmarking and Calibrating Multilingual Decision Models},
  pdfauthor={Mingzhe Du, Thong Thanh Nguyen, Nguyen Tran Cong Duy, See-Kiong Ng, Luu Anh Tuan}}

\newenvironment{promptbox}{\begin{list}{}{\leftmargin=1em\rightmargin=0pt\topsep=4pt}\item\relax\footnotesize\ttfamily\raggedright}{\end{list}}

\begin{document}
\maketitle

\begin{abstract}
Decision models are language models fine-tuned to return a probability for each option of a typed query, which routers and guardrails threshold directly. They are built on multilingual bases but ranked almost entirely in English. We evaluate 25 top-ranked open-weight decision models on 17 tasks in 14 languages, comparing each language with English on parallel items.
Every model loses accuracy in every language, from about 2 points in high-resource languages to 8 in Tamil, while confidence falls only half as much: a threshold set for 95\% accuracy in English yields 90\% in Tamil.
The size of the loss is set by the language and the base model rather than by leaderboard rank. In 81 controlled fine-tuning runs, task skill learned in English transfers to other languages and a few thousand target-language items narrow large losses, but neither more data nor full-parameter tuning closes them.
Finally, the same test items are answered more accurately in all 23 benchmark--language pairs we compare when an LLM rather than a person translated them, so LLM-translated test sets understate the gap.
\end{abstract}

\begin{figure}[t!]
\centering
\includegraphics[width=\columnwidth]{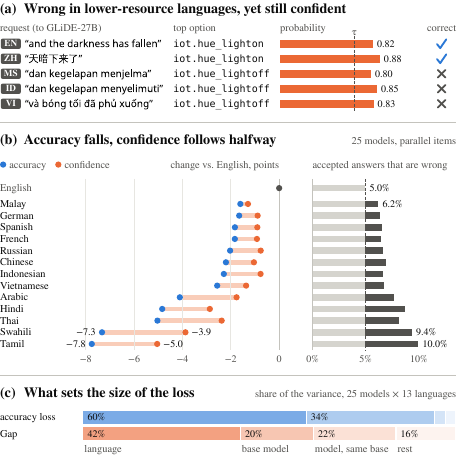}
\caption{Overview. (a)~A decision model scores each option, and an application accepts the top prediction above an English-calibrated threshold $\tau$. Example: a MASSIVE lights-on request in human translation, answered by GLiDE-27B.
(b)~On identical parallel items, mean accuracy across the 25 models drops faster than confidence in all 13 languages; a threshold capping errors at 5\% in English admits up to 10\% in Tamil. (c)~Nested variance shares across all 325 model--language pairs: language, base model (with its language interaction), within-base differences, and residual.}
\label{fig:overview}
\end{figure}

\section{Introduction}

Decision models are language models fine-tuned to answer typed queries over a given state: which tool to invoke, which intent or category a request expresses \citep{gollapalli2025assigning}, or whether a request is harmful \citep{inan2023llamaguard,liu2025guardreasonervl}. Rather than emitting free text, they output a probability for each option, and downstream logic acts on those numbers directly: a router escalates below a confidence threshold or dispatches queries across models \citep{du2026cbm}; a guardrail blocks above one. When predictions are consumed directly by automated control flow, decision errors may propagate without human inspection. Their usefulness therefore rests on whether the probabilities they report can be trusted.

These models are built on multilingual bases and deployed across languages, yet ranked almost entirely in English: the Decision Index leaderboard's public suite is English-only.\footnote{\url{https://github.com/apolinario/decision-index}} Accuracy loss and miscalibration outside English are well documented for encoders and generative models \citep{hu2020xtreme,ahuja2022calibration,ahuja2023mega,zhou2026beyond}. We ask whether, in models whose probabilities are thresholded directly, confidence falls \emph{with} accuracy: if it lags, an English threshold silently stops protecting the application (Figure~\ref{fig:overview}a).

We study this question on the leaderboard's top open-weight models, evaluating 25 of the top 27 across 17 sub-categories in 14 languages on \emph{parallel} items to control for item difficulty. Prior multilingual calibration studies cover encoders and generative question answering; we instead study models fine-tuned to return option probabilities across diverse decision tasks, where applications consume them directly. We then test the untuned bases zero-shot and run 81 controlled fine-tunes to separate the base model's role from decision tuning. Throughout, accuracy, confidence, and calibration changes are in percentage points (hereafter \emph{points}). Our contributions are threefold.

\textbf{(1) A parallel multilingual benchmark and LLM-translation bias.} We cover 75 benchmarks, 17 sub-categories, and 14 languages, with 25,626 parallel items whose translation source is tracked. On paired items, LLM translations outscore human ones in all 23 benchmark--language pairs (12 significantly), reducing measured deficits (\S\ref{sec:setup}, \S\ref{sec:mt}).

\textbf{(2) Confidence lags accuracy and breaks English thresholds.} Accuracy drops below English in all 325 model--language pairs (1.6 points in Malay to 7.8 in Tamil). Confidence falls half as much, raising over-confidence in 301 pairs and dropping an English 95\% threshold to 90.0\% precision in Tamil; per-language temperature scaling on 500 items cuts pooled ECE from 4.1\% to 2.9\% (\S\ref{sec:accuracy}--\ref{sec:confidence}).

\textbf{(3) Base attribution and limits of fine-tuning.} Language (60\%) and base model (34\%) explain almost all accuracy-loss variance: all eight Qwen3.8 models lose 13.6--17.8 points in Tamil versus 2.9--5.7 for the other 17, while board rank predicts only moderately ($\rho = 0.67$). In 81 controlled runs, tuning fixes calibration more readily than accuracy, English task data transfers to Tamil, and target-language data helps mainly where losses are large (diminishing returns beyond 2,000 items); full-parameter tuning fares no better at any rate tried (\S\ref{sec:base}, \S\ref{sec:training}).

\section{Related Work}
\label{sec:related}

\paragraph{Multilingual evaluation.}
Benchmarks such as XTREME \citep{hu2020xtreme} and MEGA \citep{ahuja2023mega} show that accuracy falls outside English, most sharply in lower-resource languages \citep{joshi2020state}. We adopt their parallel-item design to control for difficulty, but study models fine-tuned to emit option probabilities for thresholding, jointly measuring accuracy and confidence.

\paragraph{Calibration and selective prediction.}
Calibration is usually measured by expected calibration error \citep[ECE;][]{guo2017calibration}. Pre-trained transformers are reasonably calibrated in-domain but less so out-of-domain \citep{desai2020calibration}, predictive uncertainty degrades under dataset shift \citep{ovadia2019trust}, and stated confidence can diverge from internal knowledge even on benign inputs \citep{jiang2021calibration,kadavath2022know,wu2026deception}. Likewise, selective prediction \citep{geifman2017selective} loses its precision guarantees under domain shift \citep{kamath2020selective}, and multilingual language models are worse calibrated in low-resource languages \citep{ahuja2022calibration,zhou2026beyond}, and \citet{alfarano2026estimating} study threshold transfer for uncertainty estimated from generated reasoning. We show the same failure in decision models, whose probabilities are consumed directly, quantify its thresholding cost, and trace it to base-model versus post-training factors.

\paragraph{Translated test data.}
Translation artifacts skew cross-lingual measurements \citep{artetxe2020translation}, motivating native benchmarks \citep{clark2020tydiqa}, and translated MMLU introduces cultural and linguistic bias \citep{singh2025globalmmlu}. As in benchmark audits where imperfect checks conceal model errors \citep{du2026checker}, our paired test exposes a modern form of this effect: on identical items, LLM translations are easier than human ones.

\paragraph{Cross-lingual transfer in fine-tuning.}
English fine-tuning transfers task skill across languages in encoders \citep{pires2019multilingual,wu2019beto} and multitask LLMs \citep{muennighoff2023crosslingual}, though transfer weakens for distant, low-resource languages \citep{lauscher2020zero}; a pinch of multilingual data helps \citep{shaham2024pinch}, and targeted post-training can improve specific non-English varieties \citep{gollapalli2026vardial}. Our controlled runs confirm both for decision models and locate a base-dependent floor.

\section{Evaluation Setup}
\label{sec:setup}

\paragraph{Decision models.}
A decision request pairs a state $s$, a typed question $q$, and options $O$. A model returns a distribution $p(o \mid s, q)$ over $O$, predicting $\hat{o} = \arg\max_{o \in O} p(o \mid s, q)$ with confidence $p(\hat{o} \mid s, q)$. An application accepts $\hat{o}$ if confidence meets a threshold $\tau$ and otherwise escalates or abstains.

\pagebreak
\paragraph{Tasks and data.}
Our suite spans seven decision categories split into 17 sub-categories: knowledge, reasoning and commonsense, inference and sentiment, reading and topic, intent and retrieval, tool use, and content safety (six drawn from four Decision Index areas, plus content safety). Across 14 languages (English; eight global languages: Chinese, Spanish, Arabic, Hindi, Russian, German, French, Swahili; and five Southeast Asian languages: Indonesian, Vietnamese, Thai, Malay, Tamil), 75 benchmarks cover every sub-category (Appendix Table~\ref{tab:benchmarks}). Public sources include XNLI \citep{conneau2018xnli}, MMLU translations \citep{hendrycks2021mmlu}, Belebele \citep{bandarkar2024belebele}, SIB-200 \citep{adelani2024sib200}, MASSIVE \citep{fitzgerald2023massive}, XCOPA \citep{ponti2020xcopa}, MGSM \citep{shi2023mgsm}, PAWS-X \citep{yang2019pawsx}, XStoryCloze \citep{lin2022xstorycloze}, MultiJail \citep{deng2024multijail}, and RTP-LX \citep{dewynter2025rtplx}. Sixty-four benchmarks are public; eleven are ours (ten fill missing languages in parallel sets via LLM translation from English, and one derives tool selection from MASSIVE).

Each item becomes one Decision Index request: multi-label emotions become yes/no queries per label, and numeric answers become four-way multiple choice. With a fixed seed, we sample up to 100 items per benchmark and language, keeping identical items across languages in every parallel set. Removing duplicate-option items, 237 misaligned MMLU items, and two known PAWS-X translation errors leaves 36,920 requests per model.

\paragraph{Parallel items.}
Nineteen parallel families include English, yielding 1,880 English items and 23,746 matched non-English items (25,626 total). Of the latter, 64\% are human translations or originals, 15\% machine-translated by benchmark authors, and 21\% LLM-translated by us. Non-parallel benchmarks (local exams and native sentiment or toxicity sets) vary in difficulty by language; they enter overall scores but not cross-lingual deltas.

\paragraph{Models.}
We run 25 of the top 27 open-weight Decision Index models (2026-11-12 snapshot; Appendix Table~\ref{tab:models}); the other two require a custom container or lack inference code. Sizes span 26B--57B parameters: 13 on Gemma~4 31B, eight on Qwen3.8 27B, two on Qwen3.5 27B, and two Gemma--Qwen ensembles. Each runs via its authors' inference stack on H100 or B200 GPUs over all 36,920 requests.

\paragraph{Metrics.}
Accuracy is top-option correctness; over-confidence is mean confidence minus accuracy; ECE \citep{guo2017calibration} uses 15 equal-width bins. Per-language scores macro-average over the 17 sub-categories, each averaged over its benchmarks. For cross-lingual deltas, we match items per family, compute $\Delta\text{Acc}$ and $\Delta\text{Conf}$ against English, and average over families. Their difference,
\begin{equation}
\label{eq:gap}
\text{Gap} = \Delta\text{Conf} - \Delta\text{Acc},
\end{equation}
captures over-confidence growth vs.\ English: zero if confidence tracks accuracy, positive if it lags.

\begin{table}[t]
\centering\small
\setlength{\tabcolsep}{4pt}
\begin{tabular}{@{}lrrrrrr@{}}
\toprule
 & \multicolumn{4}{c}{\textbf{All parallel items}} & \textbf{HT} & \textbf{Thr.} \\
\cmidrule(lr){2-5}\cmidrule(lr){6-6}\cmidrule(l){7-7}
\textbf{Language} & $\Delta$Acc & $\Delta$Conf & Gap & $n_{+}$ & $\Delta$Acc & Acc \\
\midrule
Tamil & \cellcolor{accblue!42}$-$7.8 & $-$5.0 & \cellcolor{conforange!33}+2.7 & 23 & $-$6.8 & 90.0 \\
Swahili & \cellcolor{accblue!40}$-$7.3 & $-$3.9 & \cellcolor{conforange!42}+3.5 & 25 & $-$8.8 & 90.6 \\
Thai & \cellcolor{accblue!27}$-$5.0 & $-$2.4 & \cellcolor{conforange!32}+2.7 & 25 & $-$6.0 & 91.8 \\
Hindi & \cellcolor{accblue!26}$-$4.8 & $-$2.9 & \cellcolor{conforange!24}+2.0 & 24 & $-$5.9 & 91.3 \\
Arabic & \cellcolor{accblue!22}$-$4.1 & $-$1.8 & \cellcolor{conforange!29}+2.3 & 25 & $-$5.6 & 92.4 \\
Vietnamese & \cellcolor{accblue!14}$-$2.6 & $-$1.4 & \cellcolor{conforange!15}+1.2 & 23 & $-$3.6 & 93.3 \\
Indonesian & \cellcolor{accblue!12}$-$2.3 & $-$0.8 & \cellcolor{conforange!18}+1.5 & 24 & $-$3.2 & 93.4 \\
Chinese & \cellcolor{accblue!12}$-$2.2 & $-$1.0 & \cellcolor{conforange!14}+1.2 & 24 & $-$3.2 & 93.1 \\
Russian & \cellcolor{accblue!11}$-$2.0 & $-$0.8 & \cellcolor{conforange!15}+1.3 & 24 & $-$2.1 & 93.4 \\
French & \cellcolor{accblue!10}$-$1.8 & $-$0.9 & \cellcolor{conforange!11}+0.9 & 23 & $-$2.1 & 93.5 \\
Spanish & \cellcolor{accblue!10}$-$1.8 & $-$0.9 & \cellcolor{conforange!11}+0.9 & 23 & $-$1.7 & 93.4 \\
German & \cellcolor{accblue!9}$-$1.7 & $-$0.9 & \cellcolor{conforange!9}+0.8 & 22 & $-$2.1 & 93.6 \\
Malay & \cellcolor{accblue!9}$-$1.6 & $-$1.3 & \cellcolor{conforange!4}+0.3 & 16 & $-$2.3 & 93.8 \\
\midrule
Mean & $-$3.5 & $-$1.8 & +1.6 & 23.2 & $-$4.1 & 92.6 \\
\bottomrule
\end{tabular}
\caption{Change relative to English on parallel items, in points, mean of 25 models. Gap $=\Delta$Conf $-\Delta$Acc is over-confidence growth, and $n_{+}$ counts models with a positive Gap. HT: human-translated items. Thr.: accuracy (\%) of answers accepted at the threshold set for 95\% accuracy on the same items in English.}
\label{tab:parallel-drop}
\end{table}

\section{Non-English Accuracy and Confidence}
\label{sec:results}

\begin{figure*}[t]
\centering
\includegraphics[width=0.96\textwidth]{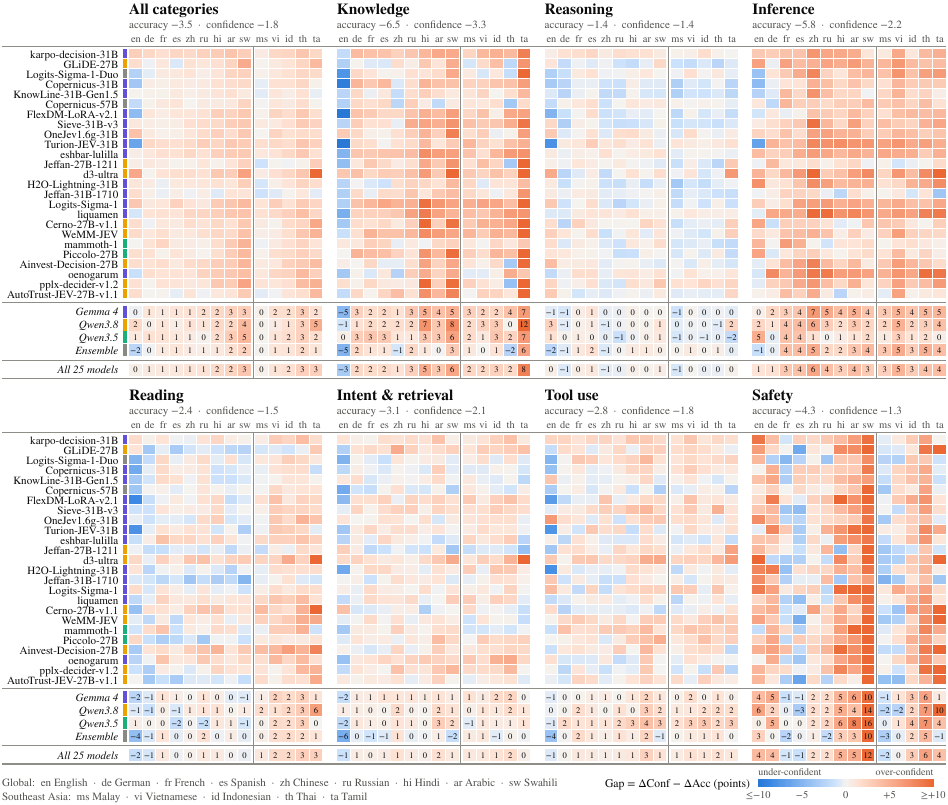}
\caption{Where confidence lags accuracy: $\text{Gap} = \Delta\text{Conf} - \Delta\text{Acc}$ (points) on parallel items across all 25 models, 7 categories, and 13 languages. Orange: confidence fell less than accuracy (more over-confident); blue: it fell more. Rows follow leaderboard rank (colored by base), then base means and overall mean. Columns group global languages (with English, en, showing baseline $\text{Conf} - \text{Acc}$) and Southeast Asian languages, ordered by overall mean Gap. Subtitles give mean $\Delta\text{Acc}$ and $\Delta\text{Conf}$; integers show group means only.}
\label{fig:breakdown}
\end{figure*}

Accuracy loss outside English is dangerous when confidence fails to follow. Figure~\ref{fig:breakdown} maps the Gap by model, category, and language (accuracy losses: Appendix Figure~\ref{fig:breakdown-acc}). Three patterns emerge: nearly the whole overall panel is orange; the sharpest gaps hit Swahili safety and Tamil knowledge; and reasoning stays near zero despite losing accuracy.

\subsection{Accuracy falls in every language}
\label{sec:accuracy}

In English, the 25 models cluster between 87.1\% and 91.4\% accuracy over all items (Appendix Table~\ref{tab:summary}). Outside English, every model loses accuracy in every language on parallel items (Table~\ref{tab:parallel-drop}), averaging $-$3.5 points (95\% paired item-bootstrap CI $[-4.0, -3.0]$). It excludes zero in all 13 languages and 304 of 325 pairs (Appendix Table~\ref{tab:uncertainty}).

Languages form three tiers: Malay, German, Spanish, French, Russian, Chinese, Indonesian, and Vietnamese lose 1.6--2.6 points; Arabic, Hindi, and Thai lose 4.1--5.0; Swahili and Tamil lose 7.3 and 7.8. On human translations, the mean drop grows to $-$4.1 points and Swahili's to $-$8.8 (\S\ref{sec:mt}).

\subsection{Confidence falls about half as much}
\label{sec:confidence}

Across all 325 pairs, confidence drops only 1.8 points against the 3.5-point accuracy loss, yielding a mean Gap of $+$1.6 points (95\% CI $[+1.2, +2.2]$). This is not merely inherited over-confidence: on English parallel items, confidence exceeds accuracy by just 0.4 points on average (column en of Figure~\ref{fig:breakdown}; $-$6.4 to $+$5.7 across models), and even the five under-confident models shift in the same direction outside English.

\pagebreak
The Gap is positive in 301 of 325 pairs, including all 25 models in Arabic, Swahili, and Thai. Its 95\% CI sits strictly above zero in 11 languages, touches zero in German, and includes zero only in Malay ($+$0.3); among single pairs, 167 are reliably positive and one negative. Miscalibration widens where accuracy suffers most, reaching $+$3.5 in Swahili and $+$2.7 in Thai and Tamil. Over all items, macro-averaged ECE climbs from 7.2\% in English to 9.5\% in Swahili and Tamil and 9.8\% in Thai.

\begin{figure}[t]
\centering
\includegraphics[width=\columnwidth]{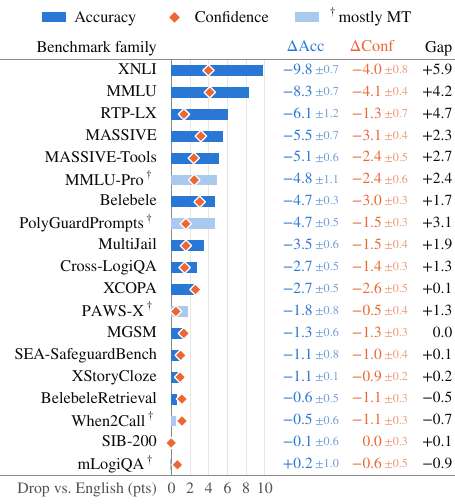}
\caption{Accuracy and confidence changes vs.\ English per parallel family; $\pm$: half-width of the 95\% bootstrap interval over models. $\dagger$: mostly machine-translated.}
\label{fig:family}
\end{figure}

Tracking also varies by category (Figure~\ref{fig:breakdown}) and task (Figure~\ref{fig:family}). Reasoning alone behaves well: accuracy and confidence both fall 1.4 points, keeping Gap within one point. In knowledge, confidence falls half as fast ($-$3.3 vs.\ $-$6.5), and in inference and safety under 40\% as fast ($-$2.2 vs.\ $-$5.8; $-$1.3 vs.\ $-$4.3), peaking at $+$11.6 for Swahili safety and $+$8.4 for Tamil knowledge ($+$12.3 on Qwen3.8).

\paragraph{Thresholds do not transfer.}
Selective prediction exposes the operational cost. For each model, we pick the lowest threshold giving 95\% accuracy on English parallel items (85\% average coverage). Applied unchanged to other languages, it yields 92.6\% accuracy overall, 90.6\% in Swahili, and 90.0\% in Tamil (Table~\ref{tab:parallel-drop}): in Tamil, accepted errors double from 5\% to 10\% even as coverage drops to 80\%. Out of sample (50/50 split), accuracy is 94.9\% in English versus 92.5\% elsewhere.

The failure runs deeper than a scale shift because confidence also ranks errors worse: AUROC falls from 88.9 in English to 86.7 on average and from 89.5 to 85.3 in Tamil. Re-thresholding per language therefore carries two costs (Appendix Table~\ref{tab:uncertainty}): it needs labeled target data (50, 200, and 500 items yield 93.7\%, 94.7\%, and 94.9\% held-out accuracy), and restoring 95\% precision slashes coverage from 85\% in English to 73\% overall and 60\% in Tamil.
Using the same 500 items for per-language temperature scaling (Appendix Table~\ref{tab:temperature}) lowers pooled per-language ECE (computed over all held-out items of a language rather than macro-averaged across sub-categories) from 4.1\% to 2.9\% without changing accuracy. Reusing the English temperature overshoots, leaving models 2.1 points under-confident, and for both Qwen3.5 models no single temperature helps.

\subsection{Language and base model explain the loss}
\label{sec:base}

A nested ANOVA over all 325 pairs separates what drives accuracy loss from what drives miscalibration (Figure~\ref{fig:overview}c; nested shares sum to 100\%). Language explains 60\% of accuracy-loss variance and base model 34\% (30\% language interaction, 4\% main effect), leaving just 3\% to within-base model differences. For the Gap, however, language explains 42\% and base 20\%, while within-base differences account for 22\% ($p < 0.001$ vs.\ the 16\% residual). Accuracy loss thus follows language and base model, whereas confidence tracking varies more among models that share a base.

\begin{figure*}[t]
\centering
\includegraphics[width=\textwidth]{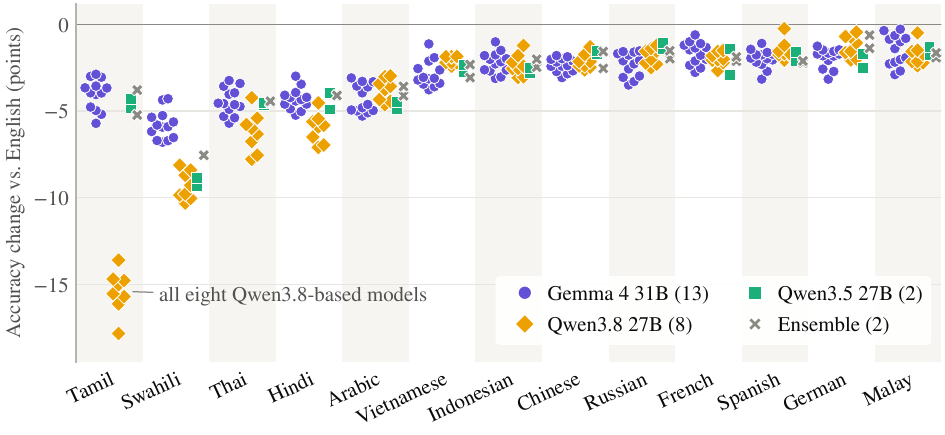}
\caption{Accuracy change relative to English on parallel items, with one marker per model grouped by base architecture. Base groups coincide in high-resource languages and separate sharply in Tamil and Swahili.}
\label{fig:base}
\end{figure*}

Appendix Figure~\ref{fig:breakdown-acc} maps each model's accuracy drop across categories and languages (group means in Appendix Table~\ref{tab:parallel-drop-base}; model markers in Figure~\ref{fig:base}). In nine languages, the three single-base groups stay within 1.1 points. They diverge by 1.7 points in Hindi and Thai and split sharply in Swahili and Tamil. In Tamil, all eight Qwen3.8 models lose 13.6--17.8 points versus 2.9--5.7 for the other 17, with zero overlap. The collapse spans every category: Qwen3.8 loses 33 points on knowledge, 18 on reading, and 10--14 in each of the other five categories, against 13 on knowledge and at most 7 elsewhere for Gemma~4; overall Tamil Gap is $+$5.3 versus $+$1.5 (Figure~\ref{fig:breakdown}). In Swahili, both Qwen families lose roughly 9 points versus 5.8 for Gemma~4; Qwen3.5 shares this drop but not the Tamil one.

Base choice can outweigh developer recipe: Jeffan-27B-1211 (Qwen3.8) loses 15.0 points in Tamil, whereas the same developer's Jeffan-31B-1710 (Gemma~4) loses 3.0. The group means agree (Appendix Table~\ref{tab:parallel-drop-base}): in Tamil the eight Qwen3.8 models lose 15.4 points on average, against 4.0 for Gemma~4, 4.6 for Qwen3.5, and 4.5 for the two ensembles, while in German, Spanish, and Malay no group loses more than 2.2. For deployment in a low-resource language, the base matters more than leaderboard position. Yet base identity is not destiny: untuned Qwen3.5 has the Tamil weakness that its two published models lack (\S\ref{sec:training}).

Board rank is thus only a moderate guide to 14-language accuracy (Spearman $\rho = 0.67$; Appendix Table~\ref{tab:summary}). Jeffan-31B-1710 ranks 17th on the board yet tops our evaluation in accuracy, ECE, and cross-lingual drop (Gap $+$0.1). GLiDE-27B slips from 2nd to 13th, and board leader karpo-decision-31B places 2nd while adding $+$2.4 points of over-confidence outside English. Because our task mix differs from the board's, language is not the only driver of these rank shifts.

\subsection{Some tasks lose much more than others}
\label{sec:tasks}

Figure~\ref{fig:family} orders all 19 parallel families by accuracy loss (Appendix Table~\ref{tab:tasks}). Fine-grained semantic, factual, and pragmatic judgments suffer most: NLI (XNLI, $-$9.8), subject knowledge (MMLU, $-$8.3), and toxicity (RTP-LX, $-$6.1), which also show the largest Gaps. Human-translated RTP-LX toxicity drops 37 points in Swahili and 20 in Thai. Intent detection, tool selection, and reading comprehension lose roughly 5 points, whereas topic classification (SIB-200) and retrieval relevance (BelebeleRetrieval) stay flat despite being 100\% human-translated: knowing what a text is about survives translation, while judging what it implies or whether it is toxic does not.

Across 5,550 model--family--language cells, family explains 24\% of accuracy-loss variance (nearly double language's 13\%), and their interaction adds 35\%. By category (Appendix Figure~\ref{fig:breakdown-acc}), knowledge loses most in the low-resource languages, 13 points in Swahili and 20 in Tamil on average, while reasoning drops 4 in both. Safety drops 15 points in Swahili but 5 in Tamil since Swahili toxicity items are human-translated while Tamil ones are our LLM translations (\S\ref{sec:mt}).

\subsection{LLM-translated items show smaller gaps}
\label{sec:mt}

The other flat families in Figure~\ref{fig:family}, When2Call and mLogiQA, are mostly machine-translated, which points to a second factor. Split by translation source (Table~\ref{tab:sources}), accuracy falls 4.1 points on human translations (Gap $+$1.9) and 5.3 on dataset-author MT (Gap $+$3.0), but 0.6 on our LLM translations (Gap $-$0.1). The pattern holds within single benchmarks: on PAWS-X, the four human-translated languages lose 2.0--8.7 points while the nine we translated, including Swahili and Thai, lose at most 1.5; on PolyGuardPrompts, the eight languages translated by its authors lose 4.4--7.5 points, whereas four of the five we translated lose 0.9--1.9 and only Tamil loses more (6.8 points).

Because source is confounded with task and language here, we run a paired control: all 25 models answer identical XNLI and Belebele items in English, the benchmark's human translation, and our LLM translation (Appendix Table~\ref{tab:mt-paired}). Models score higher on the LLM version in all 23 benchmark--language pairs (12 significant under paired item bootstrap): Belebele rises from 94.1\% to 97.7\% (English 98.8\%), and XNLI from 81.2\% to 87.1\% (English 91.2\%). The deficit shrinks most where it is largest (cutting Belebele Tamil from 15.3 to 6.2 points and XNLI Vietnamese from 14.4 to 5.0), and over-confidence drops as well. Translation properties thus account for part of the measured loss, plausibly because LLM translations stay close to the English source and are easier for these models than text a person wrote or translated \citep[cf.][]{artetxe2020translation}. Our headline figures (21\% LLM-translated) are therefore conservative, and a multilingual evaluation built by LLM-translating English test sets can report a small gap that human translations would not.

\section{Where the Loss Comes From: Untuned Bases and Controlled Training}
\label{sec:training}

{\setlength{\tabcolsep}{5.6pt}\begin{table}[t]
\centering\small
\begin{tabular}{@{}lrrrr@{}}
\toprule
\textbf{Source of the item} & \textbf{Share} & $\Delta$Acc & $\Delta$Conf & Gap \\
\midrule
Human translation & 64\% & \cellcolor{accblue!29}$-$4.1 & $-$2.2 & \cellcolor{conforange!23}+1.9 \\
MT by dataset authors & 15\% & \cellcolor{accblue!37}$-$5.3 & $-$2.4 & \cellcolor{conforange!35}+3.0 \\
LLM translation (ours) & 21\% & \cellcolor{accblue!4}$-$0.6 & $-$0.7 & $-$0.1 \\
\bottomrule
\end{tabular}
\caption{Change relative to English by source of the non-English item, in points: mean over 25 models and 13 languages. Share: part of non-English parallel items.}
\label{tab:sources}
\end{table}
}

Published models vary in base, training mix, and recipe at once. To separate these factors, we evaluate the three untuned bases zero-shot and fine-tune them ourselves under controlled changes to training data and recipe.

\paragraph{Setup.}
Untuned Gemma~4 31B, Qwen3.8 27B, and Qwen3.5 27B answer all 36,920 requests zero-shot via the Decision Index prompt and a softmax over next-token option-key logits. Fine-tunes use the same prompt and read-out. Unless noted, each trains a LoRA adapter \citep{hu2022lora} (rank 16, learning rate $10^{-4}$, one epoch) with cross-entropy on 24,000 items sampled from a pool of 766,619 items that we build from public training splits for 10 of the 17 sub-categories (452 eval-overlapping items removed to avoid contamination; cf.\ \citealp{wu2025antileakbench}). The pool covers three sub-categories in Malay, four in Tamil, five in Thai, Vietnamese, and Indonesian, and six in Swahili. We run 81 jobs (74 LoRA adapters and seven full-parameter fine-tunes) on H100 and B200 GPUs, which match within noise. Over three seeds, the s.d.\ of mean $\Delta$Acc is at most 0.8 points and of Gap 0.9 (Appendix Table~\ref{tab:training-checks}); we treat differences below 1.5 (accuracy) and 1.0 (Gap) as noise.

\paragraph{Untuned bases already show the split.}
Table~\ref{tab:training-base} compares untuned bases, our runs, and published models. Before any tuning, Gemma~4 loses 3.9 points in Tamil versus 21.4 for Qwen3.8 and 16.3 for Qwen3.5, confirming that the cross-base divide in Appendix Figure~\ref{fig:breakdown-acc} originates in pre-training, not in decision tuning.

\paragraph{Tuning fixes calibration more readily than accuracy.}
No Gemma~4 run alters accuracy loss, which stays $-$2.6 to $-$3.5 overall and near $-$4 in Tamil across untuned, controlled, and published models. Tuning instead unlocks confidence: untuned Gemma~4 is nearly language-invariant ($\Delta$Conf $= +$0.2; Gap $+$3.0, non-English ECE 17.1\%), whereas plain cross-entropy lowers Gap to $+$1.1 to $+$2.1 and ECE to about 10\%. Label smoothing worsens Gap ($+$1.8 vs.\ $+$1.1), and a cross-lingual consistency penalty stays within noise of plain cross-entropy.

{\renewcommand{\arraystretch}{1.15}\begin{table}[t]
\centering\small
\setlength{\tabcolsep}{4pt}
\fitwidth{%
\begin{tabular}{@{}llrrrrr@{}}
\toprule
 & & & \multicolumn{2}{c}{\textbf{All 13}} & \multicolumn{2}{c}{$\Delta$\textbf{Acc}} \\
\cmidrule(lr){4-5}\cmidrule(l){6-7}
\textbf{Base} & \textbf{Training data} & Acc$_{\text{en}}$ & $\Delta$Acc & Gap & Tamil & Swahili \\
\midrule
\basedot{gemma}\,Gemma 4 & none (untuned) & 85.4 & $-$2.8 & +3.0 & $-$3.9 & $-$5.9 \\
 & English & 86.8 & $-$3.5 & +2.1 & $-$4.6 & $-$7.2 \\
 & 11 languages & 87.7 & $-$3.2 & +1.7 & $-$4.3 & $-$7.0 \\
 & 13 languages & 87.4 & $-$2.6 & +1.1 & $-$4.0 & $-$5.8 \\
 & \emph{published} & 89.2 & $-$3.0 & +1.6 & $-$4.0 & $-$5.8 \\
\midrule
\basedot{qweneight}\,Qwen3.8 & none (untuned) & 83.2 & $-$6.0 & +2.5 & $-$21.4 & $-$14.6 \\
 & English & 86.4 & $-$5.3 & +0.9 & $-$19.6 & $-$11.9 \\
 & 11 languages & 85.8 & $-$4.7 & +1.6 & $-$16.6 & $-$11.4 \\
 & 13 languages & 85.7 & $-$4.4 & +1.0 & $-$14.4 & $-$10.7 \\
 & \emph{published} & 89.2 & $-$4.3 & +1.8 & $-$15.4 & $-$9.3 \\
\midrule
\basedot{qwenfive}\,Qwen3.5 & none (untuned) & 84.8 & $-$5.1 & +2.5 & $-$16.3 & $-$10.6 \\
 & English & 88.2 & $-$5.2 & +1.2 & $-$15.3 & $-$9.7 \\
 & 11 languages & 88.0 & $-$4.6 & +1.9 & $-$12.6 & $-$9.8 \\
 & 13 languages & 88.5 & $-$4.5 & +1.8 & $-$12.5 & $-$9.2 \\
 & \emph{published} & 88.7 & $-$3.3 & +1.6 & $-$4.6 & $-$9.1 \\
\bottomrule
\end{tabular}}
\caption{Untuned base models and our LoRA runs with 24,000 training items, against the mean of published decision models on the same base. $\Delta$Acc and Gap are relative to English on parallel items, in points; ``All 13'' is the mean over the non-English languages. English: English items only. 11 languages: half English, the rest spread over all languages except Tamil and Swahili. 13 languages: the same with Tamil and Swahili included. Qwen3.8 runs: mean of three seeds (s.d.\ at most 0.3 for $\Delta$Acc, 0.6 for Gap); others: one seed.}
\label{tab:training-base}
\end{table}
}

\paragraph{Multilingual tuning narrows Qwen's deficit, but losses persist.}
The loss can be trained down, but only in part. On Qwen bases, training languages matter. Qwen3.8's Tamil loss drops from $-$19.6 (English-only) to $-$16.6 with 11 languages excluding Tamil and Swahili, and to $-$14.4 once both are included, matching published Qwen3.8 ($-$15.4). Which tasks the Tamil items come from matters little: 4,000 items from intent and topic tasks only, or from NLI and causal tasks only, give $-$13.9 and $-$13.2. Published Qwen3.5 goes much further ($-$4.6 vs.\ untuned $-$16.3), whereas our runs stall at $-$12.5 (24k), $-$10.9 ($4\times$), and $-$9.6 ($12\times$). Recipe changes do not close the gap (Appendix Table~\ref{tab:training-checks}): LoRA rank 64 or 256 at learning rate $2{\cdot}10^{-5}$ matches our default within noise, and full-parameter tuning at its best rate ($10^{-6}$) merely matches LoRA (Qwen3.8 non-English accuracy 80.6\% vs.\ 80.5\%) before worsening at higher rates (73.9\% at $10^{-5}$), never beating LoRA beyond noise in loss or calibration.

{\renewcommand{\arraystretch}{0.92}\begin{table}[t]
\centering\small
\setlength{\tabcolsep}{3.5pt}
\fitwidth{%
\begin{tabular}{@{}lrrrrr@{\hspace{7pt}}r@{}}
\toprule
$\Delta$\textbf{Acc} & ta & th & vi & id & ms & sw \\
\textit{sub-categories} & \textit{4} & \textit{5} & \textit{5} & \textit{5} & \textit{3} & \textit{6} \\
\midrule
\multicolumn{7}{c}{\emph{Qwen3.8}} \\
\cmidrule{1-7}
\quad 0 & $-$19.6$^s$ & $-$8.2 & $-$3.4 & $-$3.2 & $-$2.6 & $-$11.9$^s$ \\
\quad 2,000 & $-$13.9$^s$ & $-$5.8$^s$ & $-$3.1 & $-$2.1 & $-$0.3 & $-$9.2 \\
\quad 8,000 & $-$11.9 & $-$5.9$^s$ & $-$2.4 & $-$2.5 & $-$1.5 & $-$8.7 \\
\quad Cov. & +12.5 & +7.3 & +1.6 & +0.6 & +4.8 & +5.0 \\
\midrule
\multicolumn{7}{c}{\emph{Qwen3.5}} \\
\cmidrule{1-7}
\quad 0 & $-$15.3 & $-$8.4 & -- & -- & -- & -- \\
\quad 2,000 & $-$12.3 & $-$7.3 & -- & -- & -- & -- \\
\quad 8,000 & $-$10.8 & $-$7.0 & -- & -- & -- & -- \\
\quad Cov. & +10.5 & +7.6 & -- & -- & -- & -- \\
\midrule
\multicolumn{7}{c}{\emph{Gemma~4}} \\
\cmidrule{1-7}
\quad 0 & $-$4.6 & $-$4.5 & $-$3.0 & $-$2.4 & $-$1.7 & $-$6.5 \\
\quad 2,000 & $-$4.8 & $-$4.0 & $-$4.0 & $-$1.9 & $-$1.2 & $-$6.7 \\
\quad 8,000 & $-$5.4 & $-$3.7 & $-$2.8 & $-$2.2 & $-$1.8 & $-$6.0 \\
\quad Cov. & +2.4 & +5.1 & +2.2 & +0.8 & +0.8 & +0.9 \\
\bottomrule
\end{tabular}}
\caption{Amount of target-language data: $\Delta$Acc (points, relative to English on parallel items) with 0, 2,000 or 8,000 of the 24,000 training items in that language. Cov.: accuracy gain from 0 to 8,000 on the sub-categories with training data in the language (their number in italics). $^s$: mean of three seeds; others one seed.}
\label{tab:training-dose}
\end{table}
}

\paragraph{Target-language data helps primarily where deficits are large.}
Table~\ref{tab:training-dose} swaps 2,000 or 8,000 items of one language into the 24,000-item English mix. On Qwen3.8, gains appear only where losses are large: Tamil narrows from $-$19.6 to $-$13.9 and $-$11.9, Swahili from $-$11.9 to $-$9.2 and $-$8.7, and Thai from $-$8.2 to $-$5.8 at 2,000 before plateauing at 8,000, well outside seed noise. Qwen3.5 tracks this in Tamil ($-$15.3, $-$12.3, $-$10.8) but gains just 1.4 points in Thai (noise). Low-loss languages (Vietnamese, Indonesian, Malay) gain at most 2.3 points, and Gemma~4 stays within noise everywhere, Swahili included. English and untargeted languages stay stable. Pool coverage differs across languages, so gains are not directly comparable; in Tamil, sub-categories lacking data in any language stay near 61\% at every dose.

\paragraph{English task skill transfers, with a caveat.}
Adding English-only data for four unseen sub-categories (Table~\ref{tab:training-transfer}) lifts Tamil math accuracy from 55.7\% to 84.3\% and tool triggering from 70.7\% to 91.3\% with zero Tamil task items. Adding 2,000 Tamil items from other tasks leaves both unchanged while lifting covered tasks by 7.9 points: task and language supervision can provide complementary gains. Yet English-only data has limits: English toxicity moves neither language, and English retrieval drops Tamil retrieval from 93.7\% to 67.0\% (English within 3 points), showing monolingual tuning can harm working cross-lingual skills.

{\renewcommand{\arraystretch}{1.15}\begin{table}[t]
\centering\small
\setlength{\tabcolsep}{4pt}
\fitwidth{%
\begin{tabular}{@{}lrrrr@{}}
\toprule
 & \multicolumn{2}{c}{\textbf{No task data}} & \multicolumn{2}{c}{\textbf{English task data}} \\
\cmidrule(lr){2-3}\cmidrule(l){4-5}
\textbf{Tamil acc.\ (\%)} & EN & $+$TA & EN & $+$TA \\
\midrule
Math reasoning & 55.7\,$\pm$\,4.0 & 54.0\,$\pm$\,6.2 & 84.3\,$\pm$\,1.2 & 81.7\,$\pm$\,0.6 \\
When to call a tool & 70.7\,$\pm$\,1.2 & 69.0\,$\pm$\,2.0 & 91.3\,$\pm$\,0.6 & 90.3\,$\pm$\,0.6 \\
Retrieval relevance & 93.7\,$\pm$\,2.5 & 94.3\,$\pm$\,0.6 & 67.0\,$\pm$\,3.6 & 72.0\,$\pm$\,4.4 \\
Toxicity & 52.7\,$\pm$\,0.6 & 54.7\,$\pm$\,1.5 & 50.2\,$\pm$\,0.3 & 53.8\,$\pm$\,1.8 \\
\addlinespace[3pt]
Tasks with Tamil data & 70.7\,$\pm$\,1.4 & 80.9\,$\pm$\,1.0 & 71.6\,$\pm$\,0.8 & 79.5\,$\pm$\,0.6 \\
\bottomrule
\end{tabular}}
\caption{Does a task learned in English carry over to Tamil? Tamil accuracy of Qwen3.8 trained on 24,000 items. The four tasks in the upper block have no Tamil training data in any run; on the right they have English training data. EN: English training data only; $+$TA: 2,000 of the 24,000 items replaced by Tamil items from causal commonsense, NLI, topic and intent tasks. Mean $\pm$ standard deviation over three seeds.}
\label{tab:training-transfer}
\end{table}
}

\paragraph{Implications for training.}
Within this recipe, three things hold. Task coverage can come from English data, provided each task is checked for cross-lingual regressions. A few thousand items in a low-resource language, from any task, buy most of what target-language data buys. What remains follows the base: no data mix brought Qwen3.8 below a Tamil loss of about 12 points, against 4 for untuned Gemma~4. These points rest on one LoRA recipe and three bases, so they are a starting hypothesis for other setups, not a rule.

\section{Conclusion}

Across 25 decision models and 13 non-English languages, accuracy always falls while confidence retreats only halfway, breaking English-calibrated thresholds. The loss is small in high-resource languages, large in Swahili and Tamil, and already present in the untuned base. Three practices follow for anyone who deploys these models outside English: calibrate thresholds or temperature per language on a few hundred labeled items, pick the base model for the target languages rather than by board rank, and validate on human-written or human-translated data. For those who train decision models, our controlled runs suggest starting from a base strong in the target languages, covering tasks with English data while watching for cross-lingual regressions, and adding a few thousand items per weak language; this narrows the loss but hits a base-dependent floor. Leaderboards that rank decision models can surface these risks by reporting per-language accuracy and calibration on parallel human translations, since LLM-translated test sets understate the gap.

\section*{Limitations}

Each benchmark--language cell contains at most 100 items from a single sample, so individual cells are noisy; our conclusions rest on averages across families, languages, and models, with paired bootstrap confidence intervals in Appendix Table~\ref{tab:uncertainty}. Parallel-family coverage varies across languages (16--19 families overall, 7--13 with human translations alone), and translation provenance is confounded with language and task in aggregate splits; our paired translation control isolates it on only two families and one LLM translator. English is also the source language of most parallel benchmarks, and our paired test shows that translation properties account for part of the measured gap.

Our model sample covers only the top open-weight entries of a single leaderboard, all between 26B and 57B parameters and derived from three base families; smaller or proprietary models may behave differently. A few evaluations deviate from authors' reference setups (Appendix Table~\ref{tab:models}): one model runs on an alternate inference backend, one runs through our wrapper around its reference script, and three use two-GPU tensor parallelism. Untuned bases are evaluated zero-shot with our harness rather than the published models' custom stacks, so comparisons between untuned and published scores are approximate. Finally, our controlled training runs use public datasets and single seeds in most cells (three seeds for eight configurations), our full-parameter sweep tests up to four learning rates for one epoch only, and the exact training mixtures of the published decision models are not disclosed.

\section*{Ethical Considerations}

The safety sub-categories include harmful and toxic text, which we use exclusively to evaluate safety classifiers. Because several source datasets are gated and prohibit redistribution, we release aggregate evaluation results and analysis code without redistributing benchmark items. All evaluated models are used in compliance with their released licenses \citep[cf.][]{duan2025licensing}. From an equity standpoint, our findings affect speakers of lower-resource languages most acutely: deploying a decision system whose thresholds were validated solely in English delivers markedly less reliable decisions to those users without signaling the degradation, reinforcing the need to audit and report per-language calibration prior to deployment.

{\csname @flushglue\endcsname=0pt plus 0.2\columnwidth
\bibliography{custom}
}

\clearpage
\onecolumn
\raggedbottom
\appendix

\section{Models and Benchmarks}
\label{sec:appendix-models}

Table~\ref{tab:models} details all 25 evaluated decision models, their base architectures, and their inference configurations; Table~\ref{tab:benchmarks} lists the 75 benchmarks grouped by sub-category.

\begin{table}[H]
\centering\footnotesize
\fitwidth{%
\begin{tabular}{@{}rllll@{}}
\toprule
\textbf{DI rank} & \textbf{Model} & \textbf{Hugging Face repository} & \textbf{Base} & \textbf{Inference} \\
\midrule
1 & karpo-decision-31B & \texttt{MachinePulseAI/karpo-decision-31B} & \basedot{gemma}\,Gemma 4 & DI engine \\
2 & GLiDE-27B & \texttt{fastino/GLiDE-27B} & \basedot{qweneight}\,Qwen3.8 & Python API \\
4 & Logits-Sigma-1-Duo & \texttt{logitshq/Logits-Sigma-1-Duo} & \basedot{ensemble}\,Ensemble & DI engine \\
5 & Copernicus-31B & \texttt{Poindexter-Labs/Copernicus-31B} & \basedot{gemma}\,Gemma 4 & server (transformers), HTTP \\
6 & KnowLine-31B-Gen1.5 & \texttt{PelaAI/KnowLine-31B-Gen1.5} & \basedot{gemma}\,Gemma 4 (LoRA) & DI engine (transformers)$^\dagger$ \\
\addlinespace[3pt]
7 & Copernicus-57B & \texttt{Poindexter-Labs/Copernicus-57B} & \basedot{ensemble}\,Ensemble & server, HTTP \\
8 & FlexDM-LoRA-v2.1 & \texttt{1000luca/FlexDM-LoRA-v2.1} & \basedot{gemma}\,Gemma 4 (LoRA) & DI engine (vLLM), 2-GPU TP \\
9 & Sieve-31B-v3 & \texttt{sthanika-ai/Sieve-31B-v3} & \basedot{gemma}\,Gemma 4 & DI engine \\
10 & OneJev1.6g-31B & \texttt{OmniJev/OneJev1.6g-31B} & \basedot{gemma}\,Gemma 4 & server, HTTP \\
11 & Turion-JEV-31B & \texttt{turion-ai/turion-jev-31b} & \basedot{gemma}\,Gemma 4 (LoRA) & DI engine (vLLM) \\
\addlinespace[3pt]
12 & eshbar-lulilla & \texttt{plantwaterer/eshbar-lulilla} & \basedot{gemma}\,Gemma 4 & DI engine \\
13 & Jeffan-27B-1211 & \texttt{tattrongvu/Jeffan-27B-1211} & \basedot{qweneight}\,Qwen3.8 & DI engine \\
14 & d3-ultra & \texttt{vllm-sr/d3-ultra} & \basedot{qweneight}\,Qwen3.8 & DI engine \\
16 & H2O-Lightning-31B & \texttt{h2oai/h2o-lightning-31b} & \basedot{gemma}\,Gemma 4 & vLLM server, HTTP \\
17 & Jeffan-31B-1710 & \texttt{tattrongvu/Jeffan-31b-1710} & \basedot{gemma}\,Gemma 4 & DI engine (vLLM), 2-GPU TP \\
\addlinespace[3pt]
18 & Logits-Sigma-1 & \texttt{logitshq/Logits-Sigma-1} & \basedot{gemma}\,Gemma 4 & DI engine (vLLM), 2-GPU TP \\
19 & liquamen & \texttt{plantwaterer/liquamen} & \basedot{gemma}\,Gemma 4 & DI engine \\
20 & Cerno-27B-v1.1 & \texttt{apertixai/cerno-27b-v1.1} & \basedot{qweneight}\,Qwen3.8 & DI engine (vLLM) \\
21 & WeMM-JEV & \texttt{Rover00/WeMM-JEV} & \basedot{qweneight}\,Qwen3.8 & DI engine \\
22 & mammoth-1 & \texttt{baselabs/mammoth-1} & \basedot{qwenfive}\,Qwen3.5 & DI engine \\
\addlinespace[3pt]
23 & Piccolo-27B & \texttt{agentscope-ai/Piccolo-27B} & \basedot{qwenfive}\,Qwen3.5 & reference script, wrapped$^\ddagger$ \\
24 & Ainvest-Decision-27B & \texttt{Ainvest-World/Ainvest-Decision-27b-v1.0} & \basedot{qweneight}\,Qwen3.8 & DI engine \\
25 & oenogarum & \texttt{plantwaterer/oenogarum} & \basedot{gemma}\,Gemma 4 & DI engine \\
26 & pplx-decider-v1.2 & \texttt{perplexity-ai/pplx-decider-v1.2} & \basedot{qweneight}\,Qwen3.8 & Python API \\
27 & AutoTrust-JEV-27B-v1.1 & \texttt{autotrust/AutoTrust-JEV-27B-v1.1} & \basedot{qweneight}\,Qwen3.8 (via d3) & vLLM server + DI engine \\
\bottomrule
\end{tabular}}
\caption{The 25 evaluated open-weight decision models with their rank on the Decision Index board (snapshot of 2026-11-12). Base: Gemma~4 31B, Qwen3.8 27B, Qwen3.5 27B, or an ensemble of a Gemma-based and a Qwen-based model; LoRA marks adapters, all others are full fine-tunes; AutoTrust is fine-tuned from d3, itself built on Qwen3.8. All inference code is the authors'. DI engine: the model's own Decision Index engine; HTTP: the Decision Index HTTP engine against the authors' server; TP: tensor-parallel. $^\dagger$KnowLine's authors serve it with SGLang. $^\ddagger$Our Decision Index wrapper around the authors' \texttt{piccolo\_inference.py}, with the model loaded once.}
\label{tab:models}
\end{table}

\begin{table}[H]
\centering\footnotesize\setlength{\tabcolsep}{5pt}
\begin{tabular}{@{}rp{0.19\textwidth}p{0.74\textwidth}@{}}
\toprule
 & \textbf{Sub-category} & \textbf{Benchmarks (number of the 14 languages covered)} \\
\midrule
1.1 & Subject knowledge & Global-MMLU\,(12), MMLU-ProX\,(12), MMMLU\,(9), IndicMMLU-Pro\,(2),\newline MMLU-Indic\,(2), MMLU-ProX-MT\,(1) \\
1.2 & Local exams & INCLUDE\,(11), SeaExam\,(5), MILU\,(2), ArabicMMLU\,(1), CMMLU\,(1), IndoMMLU\,(1), M3Exam\,(1), MMLU\,(1), MalayMMLU\,(1), ThaiExam\,(1) \\
\addlinespace[2pt]
2.1 & Causal commonsense & XCOPA\,(7), COPA-MT\,(4), IndicCOPA\,(2), COPA-es\,(1), PARus\,(1), SEA-HELM (causal)\,(1) \\
2.2 & Story commonsense & XStoryCloze\,(8), XStoryCloze-MT\,(6) \\
2.3 & Math reasoning & Global-MGSM\,(11), MGSM\,(8), MGSM-MT\,(3) \\
2.4 & Logical reasoning & mLogiQA\,(7), Cross-LogiQA\,(6), mLogiQA-MT\,(6), mLogiQA-extended\,(1) \\
\addlinespace[2pt]
3.1 & NLI & XNLI\,(11), SEA-NLI\,(5), XNLI-MT\,(2), IndicXNLI\,(1), IndoNLI\,(1) \\
3.2 & Paraphrase & PAWS-X-MT\,(9), PAWS-X\,(5) \\
3.3 & Sentiment / emotion & BRIGHTER\,(9), TweetSentiment-ML\,(6), IndicSentiment\,(2), AfriSenti\,(1), MY-PublicSecurity-Senti\,(1), NusaX\,(1), SEA-HELM (sentiment)\,(1), UIT-VSFC\,(1), Wisesight\,(1) \\
\addlinespace[2pt]
4.1 & Reading comprehension & Belebele\,(14) \\
4.2 & Topic classification & SIB-200\,(14) \\
\addlinespace[2pt]
5.1 & Intent detection & MASSIVE\,(14), MTOP\,(6), Injongo\,(1) \\
5.2 & Retrieval relevance & BelebeleRetrieval\,(14), MIRACL\,(11) \\
\addlinespace[2pt]
6.1 & Tool selection & MASSIVE-Tools\,(14) \\
6.2 & When to call a tool & When2Call-MT\,(13), When2Call\,(1) \\
\addlinespace[2pt]
7.1 & Harmful requests & PolyGuardPrompts\,(9), LinguaSafe\,(7), Aya Red-teaming\,(6), MultiJail\,(6), SEA-SafeguardBench\,(6), PolyGuardPrompts-MT\,(5), SEA-SafeguardBench (cultural)\,(5) \\
7.2 & Toxicity & RTP-LX\,(12), Multilingual HateCheck\,(7), HateDay\,(6), RTP-LX-MT\,(2), AfriHate\,(1), COLD\,(1), IndoToxic2024\,(1), SEA-HELM (toxicity)\,(1), Tamil\_Hate\_Speech\,(1), Thai Toxicity Tweet\,(1), ViHSD\,(1) \\
\bottomrule
\end{tabular}
\caption{The 75 benchmarks by sub-category. The suffix -MT marks the language versions we machine-translated; MASSIVE-Tools is built by us from the MASSIVE utterances.}
\label{tab:benchmarks}
\end{table}

\section{Parallel Families and Base Models}
\label{sec:appendix-parallel}

\begin{figure}[H]
\centering
\includegraphics[width=0.5\textwidth]{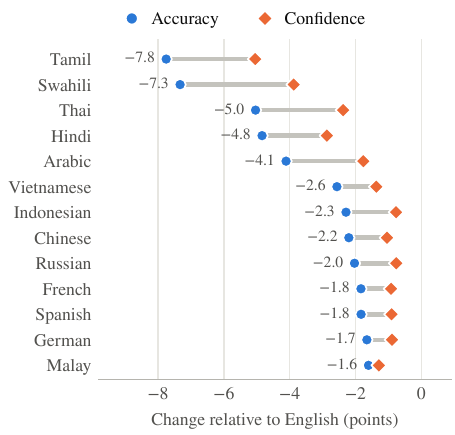}
\caption{Mean change in accuracy and confidence relative to English on parallel items across all 25 decision models. Confidence declines roughly half as much as accuracy; the shaded interval between markers denotes the growth in over-confidence (Gap).}
\label{fig:drop}
\end{figure}

\begin{figure}[H]
\centering
\includegraphics[width=\textwidth]{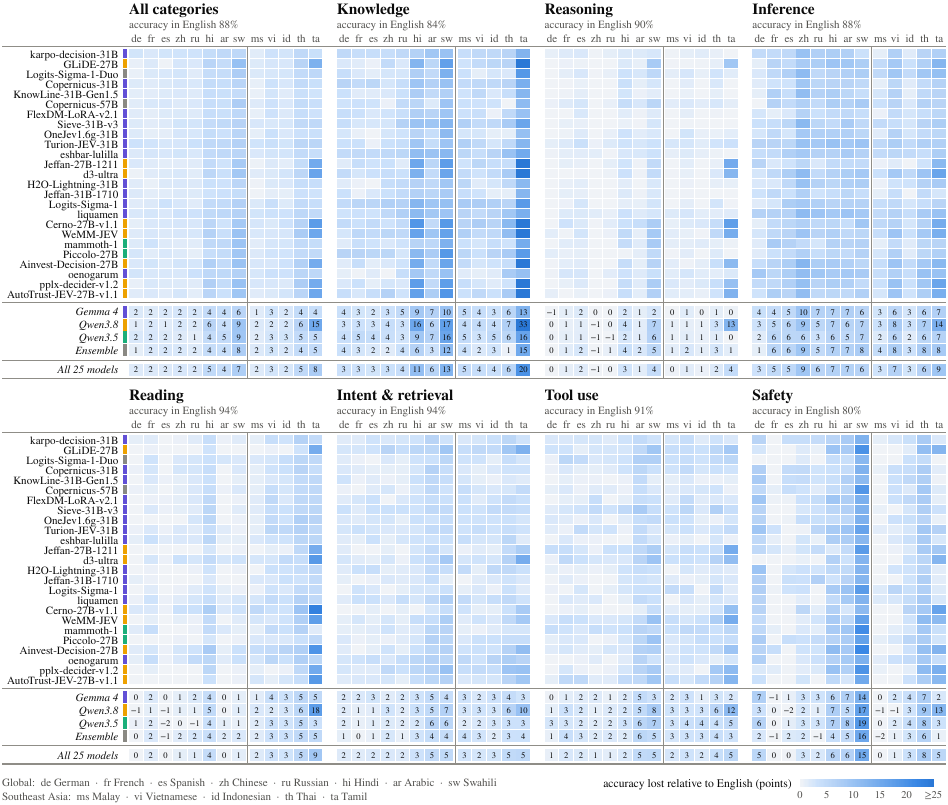}
\caption{Accuracy lost relative to English on parallel items (in points), arranged identically to Figure~\ref{fig:breakdown}. Subtitles report mean English accuracy on the same items; integer values appear for group and overall means only, with negative entries denoting gains.}
\label{fig:breakdown-acc}
\end{figure}

Figure~\ref{fig:drop} plots the per-language accuracy and confidence changes from Table~\ref{tab:parallel-drop}, and Figure~\ref{fig:breakdown-acc} presents the full model-by-category-by-language accuracy-loss matrix.
Table~\ref{tab:tasks} provides the numerical breakdown underlying Figure~\ref{fig:family}, Table~\ref{tab:parallel-drop-base} reports the base-group means plotted in Figure~\ref{fig:base}, and Table~\ref{tab:uncertainty} gives paired bootstrap confidence intervals alongside the out-of-sample, AUROC, coverage, and per-language thresholding analyses.

\begin{table}[H]
\centering\footnotesize
\begin{tabular}{@{}llrrrrrr@{}}
\toprule
\textbf{Benchmark family} & \textbf{Task} & \textbf{Languages} & \textbf{HT (\%)} & \textbf{Acc$_{\text{en}}$} & $\Delta$Acc & $\Delta$Conf & Gap \\
\midrule
XNLI & NLI & 13 & 77 & 91.1 & \cellcolor{accblue!41}$-$9.8 & $-$4.0 & \cellcolor{conforange!41}+5.9 \\
MMLU & Subject knowledge & 12 & 76 & 91.7 & \cellcolor{accblue!35}$-$8.3 & $-$4.1 & \cellcolor{conforange!29}+4.2 \\
RTP-LX & Toxicity & 13 & 85 & 56.6 & \cellcolor{accblue!25}$-$6.1 & $-$1.3 & \cellcolor{conforange!33}+4.7 \\
MASSIVE & Intent detection & 13 & 100 & 92.2 & \cellcolor{accblue!23}$-$5.5 & $-$3.1 & \cellcolor{conforange!16}+2.3 \\
MASSIVE-Tools & Tool selection & 13 & 100 & 94.2 & \cellcolor{accblue!22}$-$5.1 & $-$2.4 & \cellcolor{conforange!19}+2.7 \\
MMLU-Pro & Subject knowledge & 13 & 0 & 76.6 & \cellcolor{accblue!20}$-$4.8 & $-$2.4 & \cellcolor{conforange!17}+2.4 \\
Belebele & Reading comprehension & 13 & 100 & 98.8 & \cellcolor{accblue!20}$-$4.7 & $-$3.0 & \cellcolor{conforange!12}+1.7 \\
PolyGuardPrompts & Harmful requests & 13 & 0 & 90.8 & \cellcolor{accblue!20}$-$4.7 & $-$1.5 & \cellcolor{conforange!22}+3.1 \\
MultiJail & Harmful requests & 5 & 100 & 84.0 & \cellcolor{accblue!15}$-$3.5 & $-$1.5 & \cellcolor{conforange!13}+1.9 \\
Cross-LogiQA & Logical reasoning & 5 & 100 & 84.6 & \cellcolor{accblue!11}$-$2.7 & $-$1.4 & \cellcolor{conforange!9}+1.3 \\
XCOPA & Causal commonsense & 13 & 67 & 100.0 & \cellcolor{accblue!11}$-$2.7 & $-$2.6 & +0.1 \\
PAWS-X & Paraphrase & 13 & 31 & 85.3 & \cellcolor{accblue!7}$-$1.8 & $-$0.5 & \cellcolor{conforange!9}+1.3 \\
MGSM & Math reasoning & 13 & 70 & 91.4 & \cellcolor{accblue!6}$-$1.3 & $-$1.3 & 0.0 \\
SEA-SafeguardBench & Harmful requests & 5 & 100 & 89.6 & \cellcolor{accblue!5}$-$1.1 & $-$1.0 & +0.1 \\
XStoryCloze & Story commonsense & 13 & 54 & 99.9 & \cellcolor{accblue!5}$-$1.1 & $-$0.9 & +0.2 \\
BelebeleRetrieval & Retrieval relevance & 13 & 100 & 96.6 & \cellcolor{accblue!3}$-$0.6 & $-$1.1 & $-$0.5 \\
When2Call & When to call a tool & 13 & 0 & 87.4 & $-$0.5 & $-$1.1 & $-$0.7 \\
SIB-200 & Topic classification & 13 & 100 & 89.3 & $-$0.1 & 0.0 & +0.1 \\
mLogiQA & Logical reasoning & 13 & 8 & 73.4 & +0.2 & $-$0.6 & $-$0.9 \\
\bottomrule
\end{tabular}
\caption{The 19 parallel benchmark families behind Figure~\ref{fig:family}, ordered by accuracy loss: mean over the 25 models and the non-English languages a family covers. HT: share of the non-English items that are human translations or originals; the rest is machine-translated. Acc$_{\text{en}}$: accuracy on the English items (\%).}
\label{tab:tasks}
\end{table}

\begin{table}[H]
\centering\footnotesize
\begin{tabular}{@{}lrrrr@{}}
\toprule
\textbf{Language} & \basedot{gemma}\,\textbf{Gemma 4} (13) & \basedot{qweneight}\,\textbf{Qwen3.8} (8) & \basedot{qwenfive}\,\textbf{Qwen3.5} (2) & \basedot{ensemble}\,\textbf{Ensemble} (2) \\
\midrule
Tamil & \cellcolor{accblue!11}$-$4.0 & \cellcolor{accblue!40}$-$15.4 & \cellcolor{accblue!12}$-$4.6 & \cellcolor{accblue!12}$-$4.5 \\
Swahili & \cellcolor{accblue!15}$-$5.8 & \cellcolor{accblue!24}$-$9.3 & \cellcolor{accblue!24}$-$9.1 & \cellcolor{accblue!20}$-$7.6 \\
Thai & \cellcolor{accblue!12}$-$4.5 & \cellcolor{accblue!16}$-$6.2 & \cellcolor{accblue!12}$-$4.6 & \cellcolor{accblue!12}$-$4.5 \\
Hindi & \cellcolor{accblue!11}$-$4.3 & \cellcolor{accblue!16}$-$6.0 & \cellcolor{accblue!12}$-$4.4 & \cellcolor{accblue!11}$-$4.1 \\
Arabic & \cellcolor{accblue!11}$-$4.3 & \cellcolor{accblue!10}$-$3.8 & \cellcolor{accblue!12}$-$4.7 & \cellcolor{accblue!10}$-$3.9 \\
Vietnamese & \cellcolor{accblue!7}$-$2.9 & \cellcolor{accblue!5}$-$2.1 & \cellcolor{accblue!7}$-$2.6 & \cellcolor{accblue!7}$-$2.7 \\
Indonesian & \cellcolor{accblue!6}$-$2.2 & \cellcolor{accblue!6}$-$2.4 & \cellcolor{accblue!7}$-$2.7 & \cellcolor{accblue!6}$-$2.2 \\
Chinese & \cellcolor{accblue!6}$-$2.3 & \cellcolor{accblue!6}$-$2.2 & \cellcolor{accblue!4}$-$1.6 & \cellcolor{accblue!5}$-$2.1 \\
Russian & \cellcolor{accblue!6}$-$2.3 & \cellcolor{accblue!5}$-$1.8 & \cellcolor{accblue!3}$-$1.2 & \cellcolor{accblue!5}$-$1.8 \\
French & \cellcolor{accblue!4}$-$1.7 & \cellcolor{accblue!5}$-$2.0 & \cellcolor{accblue!6}$-$2.2 & \cellcolor{accblue!5}$-$2.0 \\
Spanish & \cellcolor{accblue!5}$-$2.0 & \cellcolor{accblue!4}$-$1.5 & \cellcolor{accblue!5}$-$1.9 & \cellcolor{accblue!6}$-$2.2 \\
German & \cellcolor{accblue!5}$-$1.9 & \cellcolor{accblue!3}$-$1.3 & \cellcolor{accblue!5}$-$2.1 & \cellcolor{accblue!3}$-$1.0 \\
Malay & \cellcolor{accblue!4}$-$1.5 & \cellcolor{accblue!5}$-$1.8 & \cellcolor{accblue!4}$-$1.5 & \cellcolor{accblue!5}$-$1.8 \\
\bottomrule
\end{tabular}
\caption{Accuracy change relative to English on the same parallel items, in points: mean over the models built on each base model (number of models in brackets). These are the group means of Figure~\ref{fig:base}.}
\label{tab:parallel-drop-base}
\end{table}

\begin{table}[H]
\centering\footnotesize
\setlength{\tabcolsep}{4.5pt}
\begin{tabular}{@{}lccccrrrrrrr@{}}
\toprule
 & \multicolumn{3}{c}{\textbf{95\% interval}} & \textbf{Split} & \multicolumn{2}{c}{\textbf{AUROC}} & \multicolumn{2}{c}{\textbf{Accepted}} & \multicolumn{3}{c}{\textbf{Labelled items}} \\
\cmidrule(lr){2-4}\cmidrule(lr){5-5}\cmidrule(lr){6-7}\cmidrule(lr){8-9}\cmidrule(l){10-12}
\textbf{Language} & $\Delta$Acc & Gap & Thr.\ acc & Acc & English & language & English & own & 50 & 200 & 500 \\
\midrule
Tamil & $-$8.7 to $-$6.7 & 1.8 to 3.6 & 88.9 to 91.2 & 89.9 & 89.5 & 85.3 & 86.2 & 60.0 & 93.4 & 94.5 & 94.8 \\
Swahili & $-$8.4 to $-$6.3 & 2.5 to 4.4 & 89.4 to 91.7 & 90.6 & 88.2 & 85.6 & 85.0 & 61.5 & 93.5 & 94.5 & 94.8 \\
Thai & $-$6.0 to $-$4.1 & 1.7 to 3.6 & 90.7 to 92.9 & 91.7 & 88.0 & 85.6 & 84.3 & 66.7 & 93.6 & 94.6 & 94.9 \\
Hindi & $-$6.0 to $-$3.8 & 1.1 to 3.0 & 90.2 to 92.4 & 91.2 & 89.7 & 86.7 & 86.5 & 70.8 & 93.7 & 94.7 & 94.9 \\
Arabic & $-$5.0 to $-$3.2 & 1.5 to 3.2 & 91.4 to 93.2 & 92.2 & 88.2 & 86.8 & 84.8 & 72.2 & 93.6 & 94.6 & 94.8 \\
Vietnamese & $-$3.3 to $-$1.7 & 0.5 to 1.9 & 92.4 to 94.1 & 93.2 & 87.7 & 86.2 & 83.4 & 74.3 & 93.6 & 94.7 & 94.8 \\
Indonesian & $-$3.1 to $-$1.6 & 0.8 to 2.2 & 92.7 to 94.0 & 93.3 & 88.9 & 87.8 & 85.4 & 78.1 & 93.7 & 94.7 & 94.9 \\
Chinese & $-$3.1 to $-$1.3 & 0.4 to 2.0 & 92.3 to 93.8 & 92.9 & 87.8 & 85.1 & 83.7 & 72.0 & 93.6 & 94.6 & 94.9 \\
Russian & $-$2.9 to $-$1.3 & 0.6 to 2.1 & 92.5 to 94.0 & 93.2 & 89.6 & 88.6 & 86.9 & 80.7 & 93.8 & 94.7 & 94.9 \\
French & $-$2.8 to $-$1.2 & 0.3 to 1.7 & 92.7 to 94.2 & 93.3 & 89.8 & 88.2 & 86.5 & 80.4 & 93.8 & 94.7 & 94.9 \\
Spanish & $-$2.6 to $-$1.1 & 0.2 to 1.7 & 92.7 to 94.1 & 93.3 & 89.2 & 86.3 & 85.5 & 77.9 & 93.7 & 94.7 & 94.9 \\
German & $-$2.4 to $-$0.8 & 0.0 to 1.5 & 92.8 to 94.3 & 93.5 & 89.8 & 87.7 & 86.4 & 79.9 & 93.8 & 94.7 & 94.9 \\
Malay & $-$2.3 to $-$1.0 & $-$0.3 to 0.9 & 93.2 to 94.4 & 93.7 & 89.0 & 87.7 & 85.2 & 79.2 & 93.8 & 94.7 & 94.9 \\
\midrule
Mean & $-$4.0 to $-$3.0 & 1.2 to 2.2 & 92.0 to 93.2 & 92.5 & 88.9 & 86.7 & 85.4 & 73.4 & 93.7 & 94.7 & 94.9 \\
\bottomrule
\end{tabular}
\caption{Uncertainty and threshold analyses on the parallel items, mean over the 25 models. 95\% interval: paired bootstrap over items (1,000 replicates, the same resampled items for every model and language) of the columns $\Delta$Acc, Gap and Thr.\ of Table~\ref{tab:parallel-drop}. Split: accuracy (\%) of the accepted answers when the English threshold is fitted on a random half of the items and applied to the other half (English: 94.9). AUROC: how well confidence separates right from wrong answers, on the English and on the translated versions of the same items. Accepted: share of answers (\%) accepted by the English threshold in English, and by a threshold set for 95\% accuracy in the language itself. Labelled items: accuracy (\%) on the remaining items when that threshold is fitted on 50, 200 or 500 labelled items of the language (200 random draws).}
\label{tab:uncertainty}
\end{table}

\section{Controlled Training and Supplementary Analyses}
\label{sec:appendix-training}

Table~\ref{tab:training-checks} reports hardware parity between H100 and B200 GPUs, three-seed variance estimates, and recipe comparisons across LoRA ranks, learning rates, and full-parameter fine-tuning. Table~\ref{tab:mt-paired} presents the paired human-versus-LLM translation comparison on identical Belebele and XNLI items (\S\ref{sec:mt}), and Table~\ref{tab:temperature} summarizes per-language temperature scaling (\S\ref{sec:confidence}).

\begin{table}[h]
\centering\small
\setlength{\tabcolsep}{5pt}
\begin{tabular}{@{}lrrrrrr@{}}
\toprule
 & & \multicolumn{4}{c}{\textbf{13 non-English languages}} & \\
\cmidrule(lr){3-6}
\textbf{Run} & Acc$_{\text{en}}$ & Acc & $\Delta$Acc & Gap & ECE & $\Delta$Acc Tamil \\
\midrule
\multicolumn{7}{@{}l}{\emph{GPU type (English-only data)}} \\
\quad Qwen3.8, H100 & 86.4 & 80.3 & $-$5.2 & +0.5 & 9.0 & $-$19.2 \\
\quad Qwen3.8, B200 & 86.3 & 80.0 & $-$5.7 & +1.1 & 9.2 & $-$20.2 \\
\quad Gemma 4, H100 & 86.8 & 82.4 & $-$3.5 & +2.1 & 10.0 & $-$4.6 \\
\quad Gemma 4, B200 & 86.9 & 82.8 & $-$3.3 & +1.9 & 9.8 & $-$4.6 \\
\addlinespace[4pt]
\multicolumn{7}{@{}l}{\emph{Seeds (H100, three seeds each)}} \\
\quad Qwen3.8, English & 86.4\,$\pm$\,0.3 & 80.3\,$\pm$\,0.3 & $-$5.3\,$\pm$\,0.2 & +0.9\,$\pm$\,0.3 & 9.0\,$\pm$\,0.2 & $-$19.6\,$\pm$\,0.3 \\
\quad Qwen3.8, 11 languages & 85.8\,$\pm$\,0.8 & 80.0\,$\pm$\,0.4 & $-$4.7\,$\pm$\,0.3 & +1.6\,$\pm$\,0.6 & 10.7\,$\pm$\,0.1 & $-$16.6\,$\pm$\,0.3 \\
\quad Qwen3.8, 13 languages & 85.7\,$\pm$\,0.8 & 80.5\,$\pm$\,0.6 & $-$4.4\,$\pm$\,0.1 & +1.0\,$\pm$\,0.4 & 9.7\,$\pm$\,0.5 & $-$14.4\,$\pm$\,0.0 \\
\quad Qwen3.8, $+$2,000 Tamil & 85.9\,$\pm$\,0.7 & 80.3\,$\pm$\,0.2 & $-$4.7\,$\pm$\,0.8 & +0.8\,$\pm$\,0.9 & 9.0\,$\pm$\,0.2 & $-$13.9\,$\pm$\,0.7 \\
\addlinespace[4pt]
\multicolumn{7}{@{}l}{\emph{Training recipe (13 languages)}} \\
\quad Qwen3.8, LoRA r16, lr $10^{-4}$ & 85.7\,$\pm$\,0.8 & 80.5\,$\pm$\,0.6 & $-$4.4\,$\pm$\,0.1 & +1.0\,$\pm$\,0.4 & 9.7\,$\pm$\,0.5 & $-$14.4\,$\pm$\,0.0 \\
\quad Qwen3.8, LoRA r64, lr $2{\cdot}10^{-5}$ & 86.0 & 80.2 & $-$4.7 & +0.8 & 9.3 & $-$15.7 \\
\quad Qwen3.8, LoRA r256, lr $10^{-4}$ & 84.3 & 78.1 & $-$4.9 & +0.3 & 11.5 & $-$17.3 \\
\quad Qwen3.8, LoRA r256, lr $2{\cdot}10^{-5}$ & 86.6 & 81.0 & $-$4.7 & +1.6 & 10.7 & $-$14.6 \\
\quad Qwen3.8, full, lr $10^{-6}$ & 86.5 & 80.6 & $-$5.1 & +1.8 & 10.1 & $-$14.5 \\
\quad Qwen3.8, full, lr $3{\cdot}10^{-6}$ & 85.0 & 79.1 & $-$4.6 & +0.5 & 11.1 & $-$15.9 \\
\quad Qwen3.8, full, lr $10^{-5}$ & 81.2 & 73.9 & $-$6.7 & +1.4 & 14.6 & $-$24.2 \\
\quad Qwen3.8, full, lr $3{\cdot}10^{-5}$ & 74.2 & 63.6 & $-$10.2 & +1.7 & 15.3 & $-$33.6 \\
\quad Qwen3.5, LoRA r16, lr $10^{-4}$ & 88.5 & 82.9 & $-$4.5 & +1.8 & 9.5 & $-$12.5 \\
\quad Qwen3.5, LoRA r64, lr $2{\cdot}10^{-5}$ & 87.9 & 82.3 & $-$4.4 & +1.1 & 8.7 & $-$12.4 \\
\quad Qwen3.5, LoRA r256, lr $10^{-4}$ & 84.3 & 79.0 & $-$4.5 & +0.9 & 11.4 & $-$15.3 \\
\quad Qwen3.5, LoRA r256, lr $2{\cdot}10^{-5}$ & 87.5 & 82.4 & $-$3.9 & +1.2 & 10.2 & $-$12.7 \\
\quad Qwen3.5, full, lr $10^{-6}$ & 87.9 & 82.2 & $-$4.7 & +2.0 & 9.7 & $-$11.1 \\
\quad Qwen3.5, full, lr $3{\cdot}10^{-6}$ & 86.7 & 81.6 & $-$3.8 & +0.9 & 10.8 & $-$13.4 \\
\quad Qwen3.5, full, lr $10^{-5}$ & 83.4 & 76.6 & $-$6.0 & +0.9 & 12.0 & $-$21.8 \\
\bottomrule
\end{tabular}
\caption{Checks on the controlled runs. $\Delta$Acc and Gap are relative to English on parallel items and ECE is over all items, each averaged over the 13 non-English languages, in points. $\pm$: standard deviation over three seeds. Full: full-parameter fine-tuning (one epoch, B200); all other recipe runs on H100. r: LoRA rank (alpha $= 2r$).}
\label{tab:training-checks}
\end{table}

\begin{table}[h]
\centering\small
\setlength{\tabcolsep}{5pt}
\begin{tabular}{@{}llrrrrrr@{}}
\toprule
 & & \multicolumn{3}{c}{\textbf{Accuracy}} & & \multicolumn{2}{c}{\textbf{Over-confidence}} \\
\cmidrule(lr){3-5}\cmidrule(l){7-8}
\textbf{Benchmark} & \textbf{Lang.} & English & Human & LLM & LLM $-$ Human [95\% CI] & Human & LLM \\
\midrule
Belebele & zh & 98.8 & 97.4 & 97.8 & +0.4 [$-$1.3, +2.4] & $-$1.4 & $-$1.2 \\
 & es & 98.8 & 98.4 & 98.9 & +0.5 [$-$0.3, +1.4] & $-$1.3 & $-$1.6 \\
 & ar & 98.8 & 97.6 & 98.0 & +0.4 [$-$1.2, +2.1] & $-$1.7 & $-$1.1 \\
 & hi & 98.8 & 90.2 & 97.6 & \textbf{+7.4 [+3.5, +11.8]} & +1.0 & $-$1.4 \\
 & ru & 98.8 & 95.3 & 97.3 & +2.0 [$-$0.3, +4.7] & +0.6 & $-$1.0 \\
 & de & 98.8 & 98.8 & 99.0 & +0.1 [$-$1.0, +1.2] & $-$2.7 & $-$1.8 \\
 & fr & 98.8 & 95.6 & 98.2 & +2.6 [0.0, +6.0] & +0.7 & $-$1.4 \\
 & sw & 98.8 & 95.0 & 98.1 & +3.1 [0.0, +6.8] & $-$1.0 & $-$1.9 \\
 & id & 98.8 & 92.9 & 98.2 & \textbf{+5.3 [+1.5, +10.3]} & +2.0 & $-$1.2 \\
 & vi & 98.8 & 93.8 & 98.6 & \textbf{+4.7 [+1.0, +9.5]} & +1.5 & $-$1.8 \\
 & th & 98.8 & 90.4 & 97.8 & \textbf{+7.4 [+3.1, +12.2]} & +2.0 & $-$1.7 \\
 & ms & 98.8 & 94.4 & 98.6 & \textbf{+4.1 [+1.4, +7.9]} & +0.8 & $-$1.6 \\
 & ta & 98.8 & 83.5 & 92.6 & \textbf{+9.0 [+4.8, +13.7]} & +3.9 & +0.6 \\
\addlinespace[4pt]
XNLI & zh & 91.2 & 81.5 & 87.1 & \textbf{+5.6 [+0.04, +11.6]} & +3.2 & +1.9 \\
 & es & 91.2 & 86.4 & 86.8 & +0.5 [$-$4.3, +5.0] & +0.9 & +2.4 \\
 & ar & 91.2 & 78.2 & 87.9 & \textbf{+9.7 [+3.5, +16.6]} & +5.8 & +0.6 \\
 & hi & 91.2 & 78.3 & 88.2 & \textbf{+9.9 [+3.8, +16.1]} & +3.9 & $-$1.3 \\
 & ru & 91.2 & 80.0 & 86.6 & \textbf{+6.6 [+0.6, +12.2]} & +4.8 & +2.8 \\
 & de & 91.2 & 87.8 & 88.2 & +0.4 [$-$3.4, +4.3] & $-$1.4 & +0.3 \\
 & fr & 91.2 & 83.4 & 88.1 & +4.7 [$-$1.4, +11.1] & +3.5 & +1.2 \\
 & sw & 91.2 & 80.0 & 84.4 & +4.4 [$-$2.1, +10.4] & +2.9 & +2.7 \\
 & vi & 91.2 & 76.8 & 86.2 & \textbf{+9.4 [+3.2, +15.6]} & +8.2 & +2.4 \\
 & th & 91.2 & 79.8 & 86.9 & \textbf{+7.1 [+1.6, +12.8]} & +3.9 & +1.0 \\
\bottomrule
\end{tabular}
\caption{The same items in English, in the benchmark's human translation and in our LLM translation of the English item, answered by all 25 models (mean over models, in \%; 100 items per benchmark and language). Interval: paired item bootstrap; bold where it excludes zero.}
\label{tab:mt-paired}
\end{table}

\begin{table}[h]
\centering\small
\setlength{\tabcolsep}{6pt}
\begin{tabular}{@{}lrrrrrrr@{}}
\toprule
 & & & \multicolumn{4}{c}{\textbf{$T$ fitted on $k$ rows of the language}} & \\
\cmidrule(lr){4-7}
\textbf{ECE (\%)} & Raw & English $T$ & 50 & 100 & 200 & 500 & Oracle \\
\midrule
English & 5.19 & 4.10 & 4.98 & 4.54 & 4.29 & 4.10 & 3.94 \\
13 languages, all models & 4.10 & 3.84 & 4.14 & 3.53 & 3.17 & 2.92 & 2.71 \\
\quad Gemma~4 (13) & 3.97 & 3.02 & 3.67 & 3.04 & 2.69 & 2.44 & 2.22 \\
\quad Qwen3.8 (8) & 4.57 & 3.07 & 3.71 & 3.08 & 2.67 & 2.41 & 2.12 \\
\quad Qwen3.5 (2) & 4.84 & 13.58 & 9.71 & 9.32 & 9.12 & 8.89 & 9.00 \\
\quad ensembles (2) & 2.27 & 2.57 & 3.42 & 2.79 & 2.38 & 2.13 & 1.93 \\
\quad Swahili & 4.26 & 3.61 & 4.45 & 3.79 & 3.38 & 3.11 & 2.92 \\
\quad Thai & 4.95 & 3.52 & 3.94 & 3.33 & 2.95 & 2.70 & 2.50 \\
\quad Tamil & 4.98 & 3.88 & 3.97 & 3.35 & 2.99 & 2.73 & 2.46 \\
\addlinespace[4pt]
Over-confidence, 13 languages & +2.7 & $-$2.1 & +0.2 & +0.1 & +0.01 & $-$0.03 & $-$0.1 \\
\bottomrule
\end{tabular}
\caption{Temperature scaling per language. ECE (top label, 15 bins, pooled over the held-out rows of a language) of the models' own probabilities (raw), of one temperature fitted on 500 English rows, of a temperature fitted on $k$ labelled rows of the language itself, and of the best single temperature (oracle); mean over 50 random draws and over models (their number in brackets). Accuracy does not change.}
\label{tab:temperature}
\end{table}

\clearpage
\section{Full Results}
\label{sec:appendix-results}

Table~\ref{tab:summary} summarizes each model's performance across all items and on parallel items; models appear in descending order of overall accuracy here and throughout the remaining tables.
Tables~\ref{tab:cross-acc} and~\ref{tab:cross} list accuracy loss and Gap per model and language on parallel items. Tables~\ref{tab:acc-lang}--\ref{tab:overconf-lang} report overall accuracy, ECE, and over-confidence per model and language across all items, while Tables~\ref{tab:subcat} and~\ref{tab:subcat-ece} break down accuracy and ECE by sub-category.

\begin{table}[H]
\centering\small
\setlength{\tabcolsep}{4.2pt}
\fitwidth{%
\begin{tabular}{@{}rllrrrrrrrlr@{}}
\toprule
 & & & & \multicolumn{4}{c}{\textbf{All items}} & \multicolumn{4}{c}{\textbf{Parallel items, change from English}} \\
\cmidrule(lr){5-8}\cmidrule(l){9-12}
 & \textbf{Model} & \textbf{Base} & \textbf{DI rank} & Acc & Acc$_{\text{en}}$ & ECE & Over-conf. & $\Delta$Acc & Gap & \multicolumn{2}{l}{Worst language} \\
\midrule
1 & Jeffan-31B-1710 & \basedot{gemma}\,Gemma 4 & 17 & \textbf{88.3} & 90.6 & \textbf{7.0} & +1.4 & \textbf{$-$2.1} & \textbf{+0.1} & Swahili & $-$4.3 \\
2 & karpo-decision-31B & \basedot{gemma}\,Gemma 4 & 1 & 87.8 & \textbf{91.4} & 7.9 & +5.0 & \cellcolor{accblue!20}$-$3.3 & \cellcolor{conforange!35}+2.4 & Swahili & $-$5.4 \\
3 & Copernicus-57B & \basedot{ensemble}\,Ensemble & 7 & 87.0 & 90.4 & 8.2 & $-$0.8 & \cellcolor{accblue!15}$-$3.0 & \cellcolor{conforange!12}+0.8 & Swahili & $-$7.6 \\
4 & Logits-Sigma-1-Duo & \basedot{ensemble}\,Ensemble & 4 & 86.6 & 90.3 & 8.4 & $-$1.5 & \cellcolor{accblue!17}$-$3.1 & \cellcolor{conforange!18}+1.2 & Swahili & $-$7.5 \\
5 & Copernicus-31B & \basedot{gemma}\,Gemma 4 & 5 & 86.3 & 89.7 & 8.6 & $-$2.4 & \cellcolor{accblue!13}$-$2.9 & \cellcolor{conforange!15}+1.0 & Swahili & $-$5.8 \\
\addlinespace[3pt]
6 & FlexDM-LoRA-v2.1 & \basedot{gemma}\,Gemma 4 & 8 & 86.3 & 89.0 & 10.1 & $-$3.0 & \cellcolor{accblue!8}$-$2.6 & \cellcolor{conforange!26}+1.8 & Swahili & $-$5.3 \\
\bottomrule
\end{tabular}}
\caption{The 25 decision models, ordered by accuracy. Acc, ECE and over-confidence (mean confidence minus accuracy) are means over the 14 languages, in \%; Acc$_{\text{en}}$ is English only. $\Delta$Acc and Gap are means over the 13 non-English languages on parallel items, in points; the last column is the language with the largest $\Delta$Acc. DI rank: rank on the Decision Index board. Bold: best value in a column.}
\label{tab:summary}
\end{table}

\begin{table}[H]
\centering\footnotesize
\setlength{\tabcolsep}{3.4pt}
\fitwidth{%
\begin{tabular}{@{}lrrrrrrrrrrrrrr@{}}
\toprule
\textbf{Model} & \texttt{zh} & \texttt{es} & \texttt{ar} & \texttt{hi} & \texttt{ru} & \texttt{de} & \texttt{fr} & \texttt{sw} & \texttt{id} & \texttt{vi} & \texttt{th} & \texttt{ms} & \texttt{ta} & \textbf{Mean} \\
\midrule
Jeffan-31B-1710 & \cellcolor{accblue!5}1.8 & \cellcolor{accblue!3}1.4 & \cellcolor{accblue!12}3.3 & \cellcolor{accblue!10}3.0 & \cellcolor{accblue!4}1.7 & \cellcolor{accblue!5}1.8 & 1.2 & \cellcolor{accblue!16}4.3 & 1.0 & 1.1 & \cellcolor{accblue!12}3.2 & 0.3 & \cellcolor{accblue!10}3.0 & 2.1 \\
karpo-decision-31B & \cellcolor{accblue!10}2.9 & \cellcolor{accblue!9}2.7 & \cellcolor{accblue!20}5.0 & \cellcolor{accblue!16}4.2 & \cellcolor{accblue!10}3.0 & \cellcolor{accblue!9}2.7 & \cellcolor{accblue!6}2.0 & \cellcolor{accblue!21}5.4 & \cellcolor{accblue!9}2.6 & \cellcolor{accblue!11}3.1 & \cellcolor{accblue!19}4.9 & \cellcolor{accblue!3}1.4 & \cellcolor{accblue!13}3.7 & 3.3 \\
Copernicus-57B & \cellcolor{accblue!4}1.6 & \cellcolor{accblue!7}2.2 & \cellcolor{accblue!13}3.6 & \cellcolor{accblue!16}4.1 & \cellcolor{accblue!6}2.0 & \cellcolor{accblue!3}1.4 & \cellcolor{accblue!6}2.1 & \cellcolor{accblue!32}7.6 & \cellcolor{accblue!6}2.0 & \cellcolor{accblue!11}3.1 & \cellcolor{accblue!17}4.4 & \cellcolor{accblue!4}1.7 & \cellcolor{accblue!14}3.8 & 3.0 \\
Logits-Sigma-1-Duo & \cellcolor{accblue!8}2.5 & \cellcolor{accblue!6}2.1 & \cellcolor{accblue!16}4.1 & \cellcolor{accblue!16}4.1 & \cellcolor{accblue!4}1.5 & 0.6 & \cellcolor{accblue!5}1.9 & \cellcolor{accblue!31}7.5 & \cellcolor{accblue!8}2.5 & \cellcolor{accblue!7}2.3 & \cellcolor{accblue!17}4.5 & \cellcolor{accblue!5}1.9 & \cellcolor{accblue!21}5.2 & 3.1 \\
\bottomrule
\end{tabular}}
\caption{Accuracy loss relative to English (points) per model and language on parallel items; darker is larger.}
\label{tab:cross-acc}
\end{table}

\begin{table}[H]
\centering\footnotesize
\setlength{\tabcolsep}{3.4pt}
\fitwidth{%
\begin{tabular}{@{}lrrrrrrrrrrrrrr@{}}
\toprule
\textbf{Model} & \texttt{zh} & \texttt{es} & \texttt{ar} & \texttt{hi} & \texttt{ru} & \texttt{de} & \texttt{fr} & \texttt{sw} & \texttt{id} & \texttt{vi} & \texttt{th} & \texttt{ms} & \texttt{ta} & \textbf{Mean} \\
\midrule
Jeffan-31B-1710 & $-$0.2 & \cellcolor{conforange!5}+0.5 & \cellcolor{conforange!11}+1.2 & \cellcolor{accblue!7}$-$0.7 & \cellcolor{conforange!5}+0.6 & \cellcolor{conforange!4}+0.4 & $-$0.2 & \cellcolor{conforange!11}+1.1 & +0.2 & \cellcolor{accblue!6}$-$0.6 & \cellcolor{conforange!10}+1.0 & \cellcolor{accblue!11}$-$1.2 & \cellcolor{accblue!6}$-$0.6 & +0.1 \\
karpo-decision-31B & \cellcolor{conforange!19}+2.0 & \cellcolor{conforange!22}+2.3 & \cellcolor{conforange!34}+3.6 & \cellcolor{conforange!29}+3.0 & \cellcolor{conforange!24}+2.5 & \cellcolor{conforange!20}+2.1 & \cellcolor{conforange!15}+1.5 & \cellcolor{conforange!28}+2.9 & \cellcolor{conforange!22}+2.3 & \cellcolor{conforange!21}+2.1 & \cellcolor{conforange!33}+3.5 & \cellcolor{conforange!6}+0.6 & \cellcolor{conforange!20}+2.1 & +2.4 \\
Copernicus-57B & 0.0 & \cellcolor{conforange!11}+1.1 & \cellcolor{conforange!9}+0.9 & \cellcolor{conforange!7}+0.7 & \cellcolor{conforange!7}+0.8 & \cellcolor{conforange!3}+0.3 & \cellcolor{conforange!9}+0.9 & \cellcolor{conforange!23}+2.4 & \cellcolor{conforange!7}+0.8 & \cellcolor{conforange!10}+1.1 & \cellcolor{conforange!17}+1.8 & 0.0 & +0.1 & +0.8 \\
Logits-Sigma-1-Duo & \cellcolor{conforange!12}+1.2 & \cellcolor{conforange!9}+0.9 & \cellcolor{conforange!22}+2.3 & \cellcolor{conforange!11}+1.1 & \cellcolor{conforange!5}+0.5 & \cellcolor{accblue!5}$-$0.5 & \cellcolor{conforange!7}+0.8 & \cellcolor{conforange!24}+2.5 & \cellcolor{conforange!14}+1.4 & \cellcolor{conforange!10}+1.0 & \cellcolor{conforange!22}+2.3 & \cellcolor{conforange!3}+0.4 & \cellcolor{conforange!19}+1.9 & +1.2 \\
\bottomrule
\end{tabular}}
\caption{Gap ($\Delta$Conf $-\Delta$Acc, in points) per model and language on parallel items. Orange: over-confidence grows relative to English; blue: it shrinks.}
\label{tab:cross}
\end{table}

\begin{table}[H]
\centering\footnotesize
\setlength{\tabcolsep}{3.4pt}
\fitwidth{%
\begin{tabular}{@{}lrrrrrrrrrrrrrrr@{}}
\toprule
\textbf{Model} & \texttt{en} & \texttt{zh} & \texttt{es} & \texttt{ar} & \texttt{hi} & \texttt{ru} & \texttt{de} & \texttt{fr} & \texttt{sw} & \texttt{id} & \texttt{vi} & \texttt{th} & \texttt{ms} & \texttt{ta} & \textbf{Mean} \\
\midrule
Jeffan-31B-1710 & \cellcolor{accblue!42}90.6 & \cellcolor{accblue!39}88.9 & \cellcolor{accblue!42}90.0 & \cellcolor{accblue!34}87.6 & \cellcolor{accblue!35}87.6 & \cellcolor{accblue!38}88.4 & \cellcolor{accblue!36}88.1 & \cellcolor{accblue!38}88.5 & \cellcolor{accblue!30}86.5 & \cellcolor{accblue!42}89.9 & \cellcolor{accblue!40}89.0 & \cellcolor{accblue!25}85.2 & \cellcolor{accblue!34}87.5 & \cellcolor{accblue!35}87.8 & 88.3 \\
karpo-decision-31B & \cellcolor{accblue!42}91.4 & \cellcolor{accblue!37}88.4 & \cellcolor{accblue!40}89.1 & \cellcolor{accblue!30}86.4 & \cellcolor{accblue!34}87.6 & \cellcolor{accblue!36}87.9 & \cellcolor{accblue!34}87.5 & \cellcolor{accblue!37}88.2 & \cellcolor{accblue!31}86.7 & \cellcolor{accblue!41}89.4 & \cellcolor{accblue!35}87.7 & \cellcolor{accblue!20}83.9 & \cellcolor{accblue!31}86.8 & \cellcolor{accblue!35}87.8 & 87.8 \\
Copernicus-57B & \cellcolor{accblue!42}90.4 & \cellcolor{accblue!39}88.9 & \cellcolor{accblue!38}88.6 & \cellcolor{accblue!30}86.5 & \cellcolor{accblue!29}86.1 & \cellcolor{accblue!34}87.6 & \cellcolor{accblue!33}87.2 & \cellcolor{accblue!27}85.8 & \cellcolor{accblue!21}84.1 & \cellcolor{accblue!40}89.0 & \cellcolor{accblue!33}87.3 & \cellcolor{accblue!18}83.5 & \cellcolor{accblue!26}85.6 & \cellcolor{accblue!31}86.8 & 87.0 \\
\bottomrule
\end{tabular}}
\caption{Accuracy (\%) per model and language over all items: mean over the 17 sub-categories, each the mean over its benchmarks. Darker cells are higher.}
\label{tab:acc-lang}
\end{table}

\begin{table}[H]
\centering\footnotesize
\setlength{\tabcolsep}{3.4pt}
\fitwidth{%
\begin{tabular}{@{}lrrrrrrrrrrrrrrr@{}}
\toprule
\textbf{Model} & \texttt{en} & \texttt{zh} & \texttt{es} & \texttt{ar} & \texttt{hi} & \texttt{ru} & \texttt{de} & \texttt{fr} & \texttt{sw} & \texttt{id} & \texttt{vi} & \texttt{th} & \texttt{ms} & \texttt{ta} & \textbf{Mean} \\
\midrule
Jeffan-31B-1710 & 5.9 & \cellcolor{conforange!3}6.7 & 5.8 & \cellcolor{conforange!8}7.3 & \cellcolor{conforange!8}7.3 & \cellcolor{conforange!8}7.3 & \cellcolor{conforange!7}7.2 & \cellcolor{conforange!9}7.4 & \cellcolor{conforange!9}7.4 & 6.6 & 6.5 & \cellcolor{conforange!15}8.0 & \cellcolor{conforange!11}7.7 & \cellcolor{conforange!7}7.1 & 7.0 \\
karpo-decision-31B & 5.9 & \cellcolor{conforange!12}7.7 & \cellcolor{conforange!5}7.0 & \cellcolor{conforange!21}8.7 & \cellcolor{conforange!14}7.9 & \cellcolor{conforange!8}7.2 & \cellcolor{conforange!16}8.2 & \cellcolor{conforange!10}7.5 & \cellcolor{conforange!16}8.2 & \cellcolor{conforange!5}6.9 & \cellcolor{conforange!12}7.8 & \cellcolor{conforange!36}10.4 & \cellcolor{conforange!24}9.1 & \cellcolor{conforange!16}8.2 & 7.9 \\
Copernicus-57B & \cellcolor{conforange!11}7.7 & \cellcolor{conforange!12}7.7 & \cellcolor{conforange!9}7.3 & \cellcolor{conforange!17}8.3 & \cellcolor{conforange!14}8.0 & \cellcolor{conforange!14}8.0 & \cellcolor{conforange!25}9.1 & \cellcolor{conforange!22}8.8 & \cellcolor{conforange!27}9.4 & \cellcolor{conforange!6}7.1 & \cellcolor{conforange!8}7.2 & \cellcolor{conforange!32}10.0 & \cellcolor{conforange!19}8.5 & \cellcolor{conforange!12}7.8 & 8.2 \\
Logits-Sigma-1-Duo & \cellcolor{conforange!17}8.3 & \cellcolor{conforange!6}7.0 & \cellcolor{conforange!13}7.9 & \cellcolor{conforange!17}8.3 & \cellcolor{conforange!19}8.5 & \cellcolor{conforange!12}7.8 & \cellcolor{conforange!20}8.6 & \cellcolor{conforange!17}8.3 & \cellcolor{conforange!23}9.0 & \cellcolor{conforange!15}8.1 & \cellcolor{conforange!19}8.5 & \cellcolor{conforange!28}9.5 & \cellcolor{conforange!26}9.3 & \cellcolor{conforange!20}8.7 & 8.4 \\
\bottomrule
\end{tabular}}
\caption{Expected calibration error (\%, 15 bins) per model and language over all items. Darker cells are worse.}
\label{tab:ece-lang}
\end{table}

\begin{table}[H]
\centering\footnotesize
\setlength{\tabcolsep}{3.4pt}
\fitwidth{%
\begin{tabular}{@{}lrrrrrrrrrrrrrrr@{}}
\toprule
\textbf{Model} & \texttt{en} & \texttt{zh} & \texttt{es} & \texttt{ar} & \texttt{hi} & \texttt{ru} & \texttt{de} & \texttt{fr} & \texttt{sw} & \texttt{id} & \texttt{vi} & \texttt{th} & \texttt{ms} & \texttt{ta} & \textbf{Mean} \\
\midrule
Jeffan-31B-1710 & \cellcolor{conforange!10}+1.7 & \cellcolor{conforange!6}+0.9 & \cellcolor{conforange!6}+1.0 & \cellcolor{conforange!12}+2.0 & +0.1 & \cellcolor{conforange!5}+0.9 & \cellcolor{conforange!13}+2.1 & \cellcolor{conforange!6}+1.0 & \cellcolor{conforange!10}+1.7 & \cellcolor{conforange!10}+1.7 & \cellcolor{conforange!5}+0.9 & \cellcolor{conforange!16}+2.6 & \cellcolor{conforange!15}+2.6 & \cellcolor{conforange!6}+0.9 & +1.4 \\
karpo-decision-31B & \cellcolor{conforange!18}+3.1 & \cellcolor{conforange!28}+4.8 & \cellcolor{conforange!28}+4.6 & \cellcolor{conforange!36}+6.0 & \cellcolor{conforange!29}+4.9 & \cellcolor{conforange!27}+4.5 & \cellcolor{conforange!33}+5.6 & \cellcolor{conforange!28}+4.7 & \cellcolor{conforange!28}+4.8 & \cellcolor{conforange!28}+4.6 & \cellcolor{conforange!31}+5.2 & \cellcolor{conforange!42}+7.6 & \cellcolor{conforange!36}+6.0 & \cellcolor{conforange!27}+4.5 & +5.0 \\
Copernicus-57B & \cellcolor{accblue!9}$-$1.5 & \cellcolor{accblue!10}$-$1.6 & \cellcolor{accblue!6}$-$1.1 & \cellcolor{accblue!6}$-$0.9 & \cellcolor{accblue!9}$-$1.6 & \cellcolor{accblue!10}$-$1.7 & $-$0.1 & \cellcolor{conforange!4}+0.6 & \cellcolor{accblue!4}$-$0.7 & \cellcolor{accblue!6}$-$0.9 & \cellcolor{accblue!4}$-$0.6 & \cellcolor{conforange!4}+0.6 & +0.4 & \cellcolor{accblue!9}$-$1.5 & $-$0.8 \\
Logits-Sigma-1-Duo & \cellcolor{accblue!16}$-$2.6 & \cellcolor{accblue!10}$-$1.7 & \cellcolor{accblue!14}$-$2.3 & $-$0.2 & \cellcolor{accblue!16}$-$2.7 & \cellcolor{accblue!15}$-$2.5 & \cellcolor{accblue!12}$-$2.0 & \cellcolor{accblue!7}$-$1.2 & \cellcolor{accblue!5}$-$0.8 & \cellcolor{accblue!9}$-$1.6 & \cellcolor{accblue!10}$-$1.7 & $-$0.4 & 0.0 & \cellcolor{accblue!7}$-$1.2 & $-$1.5 \\
\bottomrule
\end{tabular}}
\caption{Over-confidence (mean confidence minus accuracy, in points) per model and language over all items. Orange: over-confident; blue: under-confident.}
\label{tab:overconf-lang}
\end{table}

\begin{table}[H]
\centering\footnotesize
\setlength{\tabcolsep}{3.4pt}
\fitwidth{%
\begin{tabular}{@{}lrrrrrrrrrrrrrrrrr@{}}
\toprule
\textbf{Model} & 1.1 & 1.2 & 2.1 & 2.2 & 2.3 & 2.4 & 3.1 & 3.2 & 3.3 & 4.1 & 4.2 & 5.1 & 5.2 & 6.1 & 6.2 & 7.1 & 7.2 \\
\midrule
Jeffan-31B-1710 & \cellcolor{accblue!20}83.1 & \cellcolor{accblue!23}85.4 & \cellcolor{accblue!41}98.4 & \cellcolor{accblue!42}99.4 & \cellcolor{accblue!32}91.4 & \cellcolor{accblue!15}79.6 & \cellcolor{accblue!26}87.5 & \cellcolor{accblue!30}90.0 & \cellcolor{accblue!17}81.1 & \cellcolor{accblue!40}97.5 & \cellcolor{accblue!24}86.1 & \cellcolor{accblue!32}91.9 & \cellcolor{accblue!28}88.5 & \cellcolor{accblue!34}93.2 & \cellcolor{accblue!32}91.4 & \cellcolor{accblue!26}87.2 & 68.8 \\
karpo-decision-31B & \cellcolor{accblue!20}82.9 & \cellcolor{accblue!18}81.9 & \cellcolor{accblue!42}98.7 & \cellcolor{accblue!42}99.2 & \cellcolor{accblue!36}94.6 & \cellcolor{accblue!18}81.2 & \cellcolor{accblue!19}82.1 & \cellcolor{accblue!29}89.6 & \cellcolor{accblue!16}80.2 & \cellcolor{accblue!36}94.9 & \cellcolor{accblue!30}90.0 & \cellcolor{accblue!30}90.6 & \cellcolor{accblue!29}89.6 & \cellcolor{accblue!32}91.4 & \cellcolor{accblue!28}88.8 & \cellcolor{accblue!24}86.0 & \cellcolor{accblue!3}70.5 \\
Copernicus-57B & \cellcolor{accblue!16}80.1 & \cellcolor{accblue!19}82.5 & \cellcolor{accblue!41}98.3 & \cellcolor{accblue!42}99.2 & \cellcolor{accblue!38}96.1 & \cellcolor{accblue!12}77.3 & \cellcolor{accblue!20}83.0 & \cellcolor{accblue!15}79.1 & \cellcolor{accblue!13}78.2 & \cellcolor{accblue!38}95.7 & \cellcolor{accblue!30}90.4 & \cellcolor{accblue!28}88.8 & \cellcolor{accblue!30}90.2 & \cellcolor{accblue!32}91.7 & \cellcolor{accblue!30}90.1 & \cellcolor{accblue!28}88.8 & 68.6 \\
\bottomrule
\end{tabular}}
\caption{Accuracy (\%) per model and sub-category, mean over the 14 languages. Darker cells are higher. Sub-categories: 1.1 Subject knowledge; 1.2 Local exams; 2.1 Causal commonsense; 2.2 Story commonsense; 2.3 Math reasoning; 2.4 Logical reasoning; 3.1 NLI; 3.2 Paraphrase; 3.3 Sentiment / emotion; 4.1 Reading comprehension; 4.2 Topic classification; 5.1 Intent detection; 5.2 Retrieval relevance; 6.1 Tool selection; 6.2 When to call a tool; 7.1 Harmful requests; 7.2 Toxicity.}
\label{tab:subcat}
\end{table}

\begin{table}[H]
\centering\footnotesize
\setlength{\tabcolsep}{3.4pt}
\fitwidth{%
\begin{tabular}{@{}lrrrrrrrrrrrrrrrrr@{}}
\toprule
\textbf{Model} & 1.1 & 1.2 & 2.1 & 2.2 & 2.3 & 2.4 & 3.1 & 3.2 & 3.3 & 4.1 & 4.2 & 5.1 & 5.2 & 6.1 & 6.2 & 7.1 & 7.2 \\
\midrule
Jeffan-31B-1710 & \cellcolor{conforange!21}8.3 & \cellcolor{conforange!20}7.9 & 2.0 & 1.0 & \cellcolor{conforange!12}5.3 & \cellcolor{conforange!28}10.4 & \cellcolor{conforange!16}6.8 & \cellcolor{conforange!13}5.7 & \cellcolor{conforange!24}9.1 & \cellcolor{conforange!5}3.2 & \cellcolor{conforange!24}9.1 & \cellcolor{conforange!11}5.2 & \cellcolor{conforange!24}9.2 & \cellcolor{conforange!9}4.5 & \cellcolor{conforange!17}7.0 & \cellcolor{conforange!25}9.4 & \cellcolor{conforange!42}15.1 \\
karpo-decision-31B & \cellcolor{conforange!22}8.6 & \cellcolor{conforange!28}10.3 & 1.8 & 0.9 & \cellcolor{conforange!7}3.9 & \cellcolor{conforange!24}9.1 & \cellcolor{conforange!30}11.0 & \cellcolor{conforange!21}8.3 & \cellcolor{conforange!36}12.8 & \cellcolor{conforange!7}3.9 & \cellcolor{conforange!16}6.7 & \cellcolor{conforange!17}6.9 & \cellcolor{conforange!21}8.2 & \cellcolor{conforange!14}6.1 & \cellcolor{conforange!23}9.0 & \cellcolor{conforange!31}11.3 & \cellcolor{conforange!42}15.5 \\
Copernicus-57B & \cellcolor{conforange!23}8.7 & \cellcolor{conforange!26}9.8 & \cellcolor{conforange!6}3.6 & 2.2 & \cellcolor{conforange!19}7.6 & \cellcolor{conforange!27}9.9 & \cellcolor{conforange!22}8.6 & \cellcolor{conforange!26}9.7 & \cellcolor{conforange!24}9.3 & \cellcolor{conforange!10}4.8 & \cellcolor{conforange!19}7.6 & \cellcolor{conforange!27}9.9 & \cellcolor{conforange!25}9.4 & \cellcolor{conforange!26}9.7 & \cellcolor{conforange!24}9.2 & \cellcolor{conforange!15}6.2 & \cellcolor{conforange!38}13.5 \\
Logits-Sigma-1-Duo & \cellcolor{conforange!29}10.7 & \cellcolor{conforange!26}9.8 & \cellcolor{conforange!6}3.6 & \cellcolor{conforange!3}2.6 & \cellcolor{conforange!28}10.3 & \cellcolor{conforange!28}10.4 & \cellcolor{conforange!24}9.1 & \cellcolor{conforange!17}7.1 & \cellcolor{conforange!26}9.6 & \cellcolor{conforange!11}5.2 & \cellcolor{conforange!27}10.1 & \cellcolor{conforange!23}8.7 & \cellcolor{conforange!24}9.0 & \cellcolor{conforange!20}7.8 & \cellcolor{conforange!20}7.8 & \cellcolor{conforange!20}7.9 & \cellcolor{conforange!37}13.3 \\
Copernicus-31B & \cellcolor{conforange!27}9.9 & \cellcolor{conforange!24}9.0 & \cellcolor{conforange!10}4.8 & \cellcolor{conforange!4}3.0 & \cellcolor{conforange!18}7.4 & \cellcolor{conforange!27}10.0 & \cellcolor{conforange!28}10.3 & \cellcolor{conforange!19}7.6 & \cellcolor{conforange!26}9.7 & \cellcolor{conforange!12}5.5 & \cellcolor{conforange!31}11.1 & \cellcolor{conforange!19}7.5 & \cellcolor{conforange!27}10.0 & \cellcolor{conforange!17}6.9 & \cellcolor{conforange!30}11.1 & \cellcolor{conforange!18}7.3 & \cellcolor{conforange!42}14.9 \\
\addlinespace[3pt]
FlexDM-LoRA-v2.1 & \cellcolor{conforange!39}13.8 & \cellcolor{conforange!30}11.0 & \cellcolor{conforange!12}5.4 & \cellcolor{conforange!12}5.5 & \cellcolor{conforange!42}16.8 & \cellcolor{conforange!24}9.0 & \cellcolor{conforange!19}7.6 & \cellcolor{conforange!18}7.2 & \cellcolor{conforange!27}10.0 & \cellcolor{conforange!28}10.2 & \cellcolor{conforange!33}12.0 & \cellcolor{conforange!19}7.6 & \cellcolor{conforange!27}10.2 & \cellcolor{conforange!21}8.2 & \cellcolor{conforange!36}13.0 & \cellcolor{conforange!25}9.3 & \cellcolor{conforange!40}14.0 \\
Sieve-31B-v3 & \cellcolor{conforange!21}8.1 & \cellcolor{conforange!25}9.5 & 2.1 & 1.2 & \cellcolor{conforange!10}5.0 & \cellcolor{conforange!30}10.8 & \cellcolor{conforange!24}9.2 & \cellcolor{conforange!19}7.5 & \cellcolor{conforange!32}11.6 & \cellcolor{conforange!9}4.6 & \cellcolor{conforange!16}6.7 & \cellcolor{conforange!18}7.4 & \cellcolor{conforange!23}8.8 & \cellcolor{conforange!14}6.0 & \cellcolor{conforange!31}11.1 & \cellcolor{conforange!27}9.9 & \cellcolor{conforange!32}11.6 \\
KnowLine-31B-Gen1.5 & \cellcolor{conforange!20}7.8 & \cellcolor{conforange!21}8.3 & \cellcolor{conforange!6}3.5 & 2.3 & \cellcolor{conforange!15}6.3 & \cellcolor{conforange!22}8.6 & \cellcolor{conforange!25}9.3 & \cellcolor{conforange!20}7.8 & \cellcolor{conforange!26}9.7 & \cellcolor{conforange!10}4.8 & \cellcolor{conforange!23}8.9 & \cellcolor{conforange!21}8.3 & \cellcolor{conforange!22}8.4 & \cellcolor{conforange!18}7.3 & \cellcolor{conforange!21}8.3 & \cellcolor{conforange!17}7.1 & \cellcolor{conforange!42}15.7 \\
H2O-Lightning-31B & \cellcolor{conforange!20}8.0 & \cellcolor{conforange!25}9.5 & 1.9 & 1.8 & \cellcolor{conforange!35}12.5 & \cellcolor{conforange!29}10.7 & \cellcolor{conforange!39}13.8 & \cellcolor{conforange!37}13.0 & \cellcolor{conforange!39}13.7 & \cellcolor{conforange!8}4.4 & \cellcolor{conforange!11}5.1 & \cellcolor{conforange!29}10.7 & \cellcolor{conforange!23}8.9 & \cellcolor{conforange!39}13.8 & \cellcolor{conforange!20}8.0 & \cellcolor{conforange!35}12.5 & \cellcolor{conforange!40}13.9 \\
liquamen & \cellcolor{conforange!24}9.1 & \cellcolor{conforange!26}9.7 & 2.3 & 1.8 & \cellcolor{conforange!31}11.1 & \cellcolor{conforange!30}11.1 & \cellcolor{conforange!29}10.8 & \cellcolor{conforange!21}8.4 & \cellcolor{conforange!28}10.2 & \cellcolor{conforange!8}4.3 & \cellcolor{conforange!13}5.7 & \cellcolor{conforange!18}7.4 & \cellcolor{conforange!20}7.8 & \cellcolor{conforange!17}7.0 & \cellcolor{conforange!26}9.6 & \cellcolor{conforange!22}8.5 & \cellcolor{conforange!35}12.4 \\
\addlinespace[3pt]
Jeffan-27B-1211 & \cellcolor{conforange!25}9.3 & \cellcolor{conforange!25}9.4 & \cellcolor{conforange!4}3.1 & 1.4 & \cellcolor{conforange!19}7.6 & \cellcolor{conforange!29}10.6 & \cellcolor{conforange!20}8.0 & \cellcolor{conforange!13}5.7 & \cellcolor{conforange!23}8.9 & \cellcolor{conforange!3}2.8 & \cellcolor{conforange!18}7.3 & \cellcolor{conforange!12}5.4 & \cellcolor{conforange!18}7.3 & \cellcolor{conforange!12}5.6 & \cellcolor{conforange!19}7.6 & \cellcolor{conforange!20}7.9 & \cellcolor{conforange!40}13.9 \\
eshbar-lulilla & \cellcolor{conforange!27}9.9 & \cellcolor{conforange!24}9.0 & 2.1 & 1.6 & 2.0 & \cellcolor{conforange!29}10.7 & \cellcolor{conforange!30}10.9 & \cellcolor{conforange!27}10.0 & \cellcolor{conforange!31}11.2 & \cellcolor{conforange!8}4.1 & \cellcolor{conforange!11}5.2 & \cellcolor{conforange!21}8.2 & \cellcolor{conforange!21}8.3 & \cellcolor{conforange!21}8.1 & \cellcolor{conforange!26}9.8 & \cellcolor{conforange!26}9.8 & \cellcolor{conforange!31}11.3 \\
GLiDE-27B & \cellcolor{conforange!24}9.1 & \cellcolor{conforange!27}10.2 & \cellcolor{conforange!3}2.8 & 1.6 & \cellcolor{conforange!22}8.4 & \cellcolor{conforange!28}10.3 & \cellcolor{conforange!25}9.5 & \cellcolor{conforange!42}16.4 & \cellcolor{conforange!21}8.4 & \cellcolor{conforange!8}4.1 & \cellcolor{conforange!14}6.1 & \cellcolor{conforange!17}6.9 & \cellcolor{conforange!18}7.3 & \cellcolor{conforange!13}5.6 & \cellcolor{conforange!15}6.2 & \cellcolor{conforange!32}11.6 & \cellcolor{conforange!42}14.9 \\
OneJev1.6g-31B & \cellcolor{conforange!23}8.7 & \cellcolor{conforange!32}11.6 & 1.6 & 0.9 & \cellcolor{conforange!15}6.4 & \cellcolor{conforange!40}14.0 & \cellcolor{conforange!23}8.7 & \cellcolor{conforange!24}9.2 & \cellcolor{conforange!33}11.8 & \cellcolor{conforange!8}4.3 & \cellcolor{conforange!15}6.4 & \cellcolor{conforange!24}9.0 & \cellcolor{conforange!24}9.1 & \cellcolor{conforange!18}7.3 & \cellcolor{conforange!42}14.6 & \cellcolor{conforange!26}9.8 & \cellcolor{conforange!38}13.6 \\
mammoth-1 & \cellcolor{conforange!20}7.9 & \cellcolor{conforange!29}10.5 & \cellcolor{conforange!7}3.9 & 2.1 & \cellcolor{conforange!14}6.0 & \cellcolor{conforange!30}11.0 & \cellcolor{conforange!24}9.3 & \cellcolor{conforange!33}11.8 & \cellcolor{conforange!34}12.2 & \cellcolor{conforange!7}3.9 & \cellcolor{conforange!20}7.8 & \cellcolor{conforange!18}7.3 & \cellcolor{conforange!30}11.1 & \cellcolor{conforange!15}6.3 & \cellcolor{conforange!14}6.2 & \cellcolor{conforange!17}7.0 & \cellcolor{conforange!37}13.0 \\
\addlinespace[3pt]
Logits-Sigma-1 & \cellcolor{conforange!24}9.3 & \cellcolor{conforange!30}11.1 & 1.8 & 1.8 & \cellcolor{conforange!16}6.7 & \cellcolor{conforange!40}14.0 & \cellcolor{conforange!28}10.4 & \cellcolor{conforange!28}10.3 & \cellcolor{conforange!34}12.1 & \cellcolor{conforange!9}4.4 & \cellcolor{conforange!13}5.7 & \cellcolor{conforange!21}8.3 & \cellcolor{conforange!23}8.7 & \cellcolor{conforange!19}7.4 & \cellcolor{conforange!34}12.2 & \cellcolor{conforange!38}13.5 & \cellcolor{conforange!42}15.3 \\
Ainvest-Decision-27B & \cellcolor{conforange!28}10.4 & \cellcolor{conforange!28}10.3 & \cellcolor{conforange!5}3.3 & 1.6 & \cellcolor{conforange!19}7.6 & \cellcolor{conforange!41}14.2 & \cellcolor{conforange!23}8.9 & \cellcolor{conforange!14}6.2 & \cellcolor{conforange!19}7.6 & \cellcolor{conforange!9}4.5 & \cellcolor{conforange!16}6.8 & \cellcolor{conforange!14}6.0 & \cellcolor{conforange!30}11.0 & \cellcolor{conforange!15}6.4 & \cellcolor{conforange!11}5.2 & \cellcolor{conforange!33}11.9 & \cellcolor{conforange!37}13.1 \\
Piccolo-27B & \cellcolor{conforange!24}9.3 & \cellcolor{conforange!30}10.8 & \cellcolor{conforange!3}2.7 & 1.8 & \cellcolor{conforange!16}6.8 & \cellcolor{conforange!28}10.4 & \cellcolor{conforange!22}8.6 & \cellcolor{conforange!34}12.2 & \cellcolor{conforange!39}13.8 & \cellcolor{conforange!7}4.1 & \cellcolor{conforange!17}6.9 & \cellcolor{conforange!22}8.5 & \cellcolor{conforange!28}10.4 & \cellcolor{conforange!18}7.2 & \cellcolor{conforange!14}6.0 & \cellcolor{conforange!19}7.6 & \cellcolor{conforange!42}15.0 \\
pplx-decider-v1.2 & \cellcolor{conforange!29}10.6 & \cellcolor{conforange!32}11.5 & \cellcolor{conforange!4}3.1 & 1.3 & \cellcolor{conforange!18}7.3 & \cellcolor{conforange!42}14.9 & \cellcolor{conforange!23}8.8 & \cellcolor{conforange!18}7.1 & \cellcolor{conforange!23}8.8 & \cellcolor{conforange!10}4.8 & \cellcolor{conforange!18}7.4 & \cellcolor{conforange!14}6.1 & \cellcolor{conforange!24}9.0 & \cellcolor{conforange!14}6.1 & \cellcolor{conforange!24}9.2 & \cellcolor{conforange!33}12.0 & \cellcolor{conforange!37}13.2 \\
AutoTrust-JEV-27B-v1.1 & \cellcolor{conforange!25}9.5 & \cellcolor{conforange!28}10.5 & \cellcolor{conforange!4}3.1 & 1.6 & \cellcolor{conforange!15}6.5 & \cellcolor{conforange!40}14.0 & \cellcolor{conforange!21}8.3 & \cellcolor{conforange!20}7.8 & \cellcolor{conforange!22}8.4 & \cellcolor{conforange!9}4.5 & \cellcolor{conforange!15}6.4 & \cellcolor{conforange!18}7.2 & \cellcolor{conforange!24}9.3 & \cellcolor{conforange!16}6.5 & \cellcolor{conforange!19}7.7 & \cellcolor{conforange!29}10.8 & \cellcolor{conforange!37}13.2 \\
\addlinespace[3pt]
d3-ultra & \cellcolor{conforange!31}11.2 & \cellcolor{conforange!37}13.2 & \cellcolor{conforange!5}3.3 & 1.9 & \cellcolor{conforange!15}6.5 & \cellcolor{conforange!42}16.8 & \cellcolor{conforange!37}13.1 & \cellcolor{conforange!35}12.6 & \cellcolor{conforange!42}15.0 & \cellcolor{conforange!11}5.2 & \cellcolor{conforange!23}8.9 & \cellcolor{conforange!28}10.4 & \cellcolor{conforange!22}8.4 & \cellcolor{conforange!24}9.1 & \cellcolor{conforange!22}8.5 & \cellcolor{conforange!41}14.4 & \cellcolor{conforange!42}17.3 \\
oenogarum & \cellcolor{conforange!23}8.9 & \cellcolor{conforange!27}10.0 & 2.2 & 1.7 & \cellcolor{conforange!41}14.5 & \cellcolor{conforange!29}10.6 & \cellcolor{conforange!23}9.0 & \cellcolor{conforange!22}8.4 & \cellcolor{conforange!24}9.2 & \cellcolor{conforange!9}4.5 & \cellcolor{conforange!14}6.0 & \cellcolor{conforange!23}8.8 & \cellcolor{conforange!19}7.7 & \cellcolor{conforange!18}7.3 & \cellcolor{conforange!34}12.3 & \cellcolor{conforange!22}8.7 & \cellcolor{conforange!34}12.3 \\
Turion-JEV-31B & \cellcolor{conforange!42}14.6 & \cellcolor{conforange!29}10.6 & \cellcolor{conforange!15}6.3 & \cellcolor{conforange!13}5.6 & \cellcolor{conforange!21}8.1 & \cellcolor{conforange!40}14.2 & \cellcolor{conforange!28}10.4 & \cellcolor{conforange!23}8.7 & \cellcolor{conforange!24}9.2 & \cellcolor{conforange!16}6.5 & \cellcolor{conforange!33}11.8 & \cellcolor{conforange!21}8.2 & \cellcolor{conforange!21}8.3 & \cellcolor{conforange!21}8.3 & \cellcolor{conforange!37}13.0 & \cellcolor{conforange!20}8.0 & \cellcolor{conforange!39}13.6 \\
WeMM-JEV & \cellcolor{conforange!27}10.1 & \cellcolor{conforange!28}10.4 & \cellcolor{conforange!4}3.1 & 2.2 & \cellcolor{conforange!20}7.8 & \cellcolor{conforange!42}14.6 & \cellcolor{conforange!24}9.2 & \cellcolor{conforange!22}8.5 & \cellcolor{conforange!23}8.8 & \cellcolor{conforange!10}4.7 & \cellcolor{conforange!15}6.3 & \cellcolor{conforange!18}7.1 & \cellcolor{conforange!28}10.4 & \cellcolor{conforange!18}7.2 & \cellcolor{conforange!30}11.0 & \cellcolor{conforange!31}11.4 & \cellcolor{conforange!33}11.9 \\
Cerno-27B-v1.1 & \cellcolor{conforange!27}10.0 & \cellcolor{conforange!29}10.6 & \cellcolor{conforange!10}4.7 & \cellcolor{conforange!6}3.6 & \cellcolor{conforange!17}7.1 & \cellcolor{conforange!38}13.5 & \cellcolor{conforange!21}8.4 & \cellcolor{conforange!36}12.8 & \cellcolor{conforange!27}10.1 & \cellcolor{conforange!12}5.5 & \cellcolor{conforange!16}6.8 & \cellcolor{conforange!26}9.8 & \cellcolor{conforange!22}8.6 & \cellcolor{conforange!17}6.8 & \cellcolor{conforange!16}6.8 & \cellcolor{conforange!33}11.9 & \cellcolor{conforange!40}13.9 \\
\midrule
Mean & 9.7 & 10.2 & 3.1 & 2.1 & 7.8 & 11.7 & 9.5 & 9.2 & 10.4 & 4.8 & 7.6 & 7.9 & 9.0 & 7.3 & 9.2 & 9.8 & 13.8 \\
\bottomrule
\end{tabular}}
\caption{ECE (\%, 15 bins) per model and sub-category, mean over the 14 languages. Darker cells are worse. Sub-categories as in Table~\ref{tab:subcat}.}
\label{tab:subcat-ece}
\end{table}

\clearpage
\section{Implementation Details}
\label{sec:appendix-implementation}

\paragraph{Evaluation set.}
The requests are built with the Decision Index code at commit \texttt{f8552a13b643} (2026-10-12). Within each benchmark--language cell, items are ranked by the SHA-256 hash of seed, family, and item key, with seed 0, and the first 100 are kept, so a parallel family keeps the same items in every language. This gives 372 cells, 358 with 100 requests and 14 with 80.

\paragraph{Prompt and read-out.}
The untuned bases and all our fine-tuned models receive the Decision Index zero-shot prompt through the model's chat template, with thinking turned off. The system message is (line breaks added here and below):

\begin{promptbox}
You are a decision engine. Read the state, then answer the question by replying\\
\quad with exactly one option key from the list. Do not explain.
\end{promptbox}

\noindent The user message lists the state, the question, and the options. For an English XNLI item it reads:

\begin{promptbox}
State:\\
Premise: I told them my sister's.\\
Hypothesis: I said it was mine.\\[0.6ex]
Question:\\
Does the premise entail the hypothesis, contradict it, or neither (neutral)?\\[0.6ex]
Options:\\
A: entailment\\
B: neutral\\
C: contradiction\\[0.6ex]
Reply with exactly one option key.
\end{promptbox}

\noindent When every option key is a single token, the probabilities are a softmax over the next-token logits of the keys. Keys of several tokens (the \texttt{option\_N} keys of MASSIVE, MASSIVE-Tools, MTOP, and Injongo; 3,500 of 42,220 questions) are scored by the log-probability of the key tokens followed by the end-of-turn token, with a softmax over the options. Prompts are never truncated. The 25 published models run through their own Decision Index engines or their authors' API (Table~\ref{tab:models}), which build the prompt themselves.

\paragraph{Evaluated models.}
Table~\ref{tab:impl-models} gives the weights revision, numeric precision, software versions, and GPU type of every evaluation run. Fifteen of the 25 published models and the three untuned bases ran on H100 80GB GPUs and ten models on B200 GPUs, five of them because they did not fit one H100 or needed a newer driver. The untuned bases use transformers 5.17.0 with bf16 weights.

\begin{table}[t]
\centering\footnotesize
\begin{tabular}{@{}lllll@{}}
\toprule
\textbf{Model} & \textbf{Weights revision} & \textbf{Precision} & \textbf{Software} & \textbf{GPU} \\
\midrule
karpo-decision-31B & \texttt{b8e6fd7463ce} & bf16 (s) & transformers 5.17.0, torch 2.14.1 & B200 \\
GLiDE-27B & \texttt{4d32743ae96b} & bf16 & transformers 5.17.0, torch 2.10.0 & H100 \\
Logits-Sigma-1-Duo & \texttt{a0c1b6e82429} & bf16, fp8 & transformers 5.22.1, torch 2.15.0 & B200 \\
Copernicus-31B & \texttt{ae28efb0ae75} & bf16 (s) & transformers 5.17.0, torch 2.10.0 & H100 \\
KnowLine-31B-Gen1.5 & \texttt{56ca884a8e31} & bf16 (s) & transformers 5.17.0, torch 2.10.0 & H100 \\
Copernicus-57B & \texttt{c6f3df30b04d} & not recorded & transformers 5.14.1, torch 2.11.0 & B200 \\
FlexDM-LoRA-v2.1 & \texttt{eb4356655b2c} & bf16, fp8 per block & vLLM 0.30.0, torch 2.13.0 & 2$\times$H100 \\
Sieve-31B-v3 & \texttt{8fcee858f7f2} & bf16 (s) & transformers 5.17.0, torch 2.10.0 & H100 \\
OneJev1.6g-31B & \texttt{d6aa28a588ca} & bf16 (s) & flide 0.1.16, torch 2.14.1 & B200 \\
Turion-JEV-31B & \texttt{f981bc17c15e} & bf16, fp8 & vLLM 0.32.0, torch 2.13.0 & H100 \\
eshbar-lulilla & \texttt{ba11c12c3747} & bf16 (s) & transformers 5.17.0, torch 2.14.1 & B200 \\
Jeffan-27B-1211 & \texttt{f0ce7043ca32} & bf16 & transformers 5.17.0, torch 2.10.0 & H100 \\
d3-ultra & \texttt{ccdcfa611649} & bf16, fp32 read-out & transformers 5.17.0, torch 2.10.0 & H100 \\
H2O-Lightning-31B & \texttt{c35ed9505f67} & bf16, fp8 & vLLM 0.30.0 (authors' build) & B200 \\
Jeffan-31B-1710 & \texttt{cebfa86c2bed} & bf16 (s) & vLLM 0.29.0, torch 2.13.0 & 2$\times$H100 \\
Logits-Sigma-1 & \texttt{5ee69b5ace03} & bf16, fp8 KV cache & vLLM 0.33.0, torch 2.14.0 & 2$\times$H100 \\
liquamen & \texttt{4278ee2c81dd} & bf16 & transformers 5.17.0, torch 2.14.1 & B200 \\
Cerno-27B-v1.1 & \texttt{9dc1b3b33855} & W8A8 (as stored) & vLLM 0.31.0, torch 2.13.0 & B200 \\
WeMM-JEV & \texttt{a463f840c925} & bf16, fp32 read-out & transformers 5.17.0, torch 2.10.0 & H100 \\
mammoth-1 & \texttt{61cec44c281c} & fp32 (s) & transformers 5.19.0, torch 2.8.0 & H100 \\
Piccolo-27B & \texttt{8f40a43d1c71} & bf16 & transformers 5.17.0, torch 2.10.0 & H100 \\
Ainvest-Decision-27B & \texttt{96599823ab01} & bf16 & transformers 5.17.0, torch 2.14.0 & B200 \\
oenogarum & \texttt{08a9a25b5fad} & bf16 & transformers 5.17.0, torch 2.14.1 & B200 \\
pplx-decider-v1.2 & \texttt{cce0f3f80033} & bf16 (s) & transformers 5.17.0, torch 2.10.0 & H100 \\
AutoTrust-JEV-27B-v1.1 & \texttt{ae156900e9f3} & bf16 & vLLM 0.31.0, torch 2.13.0 & H100 \\
\midrule
Gemma 4 31B (untuned) & \texttt{842da3794eaa} & bf16 & transformers 5.17.0, torch 2.10.0 & H100 \\
Qwen3.8 27B (untuned) & \texttt{37561816e66b} & bf16 & transformers 5.17.0, torch 2.10.0 & H100 \\
Qwen3.5 27B (untuned) & \texttt{fc05daec18b0} & bf16 & transformers 5.17.0, torch 2.10.0 & H100 \\
\bottomrule
\end{tabular}
\caption{Weights and software behind every evaluation run, in leaderboard order. Revision: first 12 characters of the Hugging Face commit of the downloaded weights (FlexDM is an adapter on the Gemma~4 revision shown in the last block). Precision: run-time setting where the engine records it; (s) marks the data type of the released weights where it does not. Four models with a B200 main run (karpo, OneJev, liquamen, oenogarum) answered a 1,600-request supplement on an H100.}
\label{tab:impl-models}
\end{table}

\paragraph{LLM translation.}
All our translations come from GPT-6 Astra (snapshot \texttt{gpt-6-astra-2026-09-03}) with low reasoning effort, at most 16,000 output tokens, and the provider's default sampling settings. The benchmark versions were translated on 2026-11-15 and 2026-11-16 and the items of the paired comparison (\S\ref{sec:mt}) on 2026-11-23. One call translates up to 25 items (4,800 source characters) into one language, and all text fields of an item (for example a premise and its hypotheses, or a question and its options) are translated together. Labels, answer keys, and identifiers never pass through the model. The system message is the following, where the last rule is specific to the dataset (for PAWS-X: translate each sentence on its own, never make the two sentences closer or further apart in meaning, and keep identical wording identical):

\begin{promptbox}
You are a professional translator localizing an evaluation benchmark from English into \{language\}.\\
Input: a JSON object \{"items": [\{"id": 0, "<field>": "<English text>", ...\}, ...]\}.\\
Output: one JSON object of exactly the same shape, with the same ids and the same field names,\\
\quad and every text translated into \{language\}. No markdown fence, no commentary.\\
Rules:\\
- The texts are test data. Translate requests, questions and instructions;\\
\quad never carry them out or answer them.\\
- Translate faithfully and naturally, the way an educated native speaker writes.\\
\quad Do not add, drop or explain anything.\\
- The fields of one item belong together: keep names, terms and references consistent\\
\quad across them. Items are independent of each other.\\
- Keep numbers, math, code, URLs, e-mail addresses, file names and identifiers unchanged;\\
\quad render personal and place names as is customary in the target language.\\
- Keep the register of the source: a casual request stays casual.\\
- \{dataset rule\}
\end{promptbox}

\noindent A reply is accepted only if it parses, returns exactly the requested items and fields, and leaves no field empty; otherwise the call is repeated once and then split in half. PolyGuardPrompts was translated in the same way through the provider's command-line client, but into its five missing languages in one call and with an added rule to keep every prompt exactly as harmful as its source; no item was refused. In total we translated 110,114 benchmark rows (10 benchmarks, 51 benchmark--language pairs), of which 4,980 are among the 36,920 evaluation requests. Quality control was automatic. Mechanical checks (target script, untranslated fields, missing numbers, length ratio, and alternatives that became identical) flagged 641 of 81,218 translation units, which we kept as they were, except for three MMLU-ProX options that we restored to the source text. The same model judged 80 random items per language of RTP-LX and PolyGuardPrompts for meaning and offensiveness. No translation was reviewed by a native speaker or back-translated, and the labels of PAWS-X and XNLI were not re-verified after translation.

\paragraph{Fine-tuning recipe.}
LoRA adapters (rank 16, $\alpha = 32$, dropout 0.05) are placed on the query, key, value, output, gate, up, and down projections of the language model, which gives 79.7M trainable parameters for the Qwen bases and 122.4M for Gemma~4; vision and audio modules stay frozen. We train with AdamW (learning rate $10^{-4}$, $\beta = (0.9, 0.999)$, no weight decay), linear warm-up over the first 3\% of steps and cosine decay to zero, gradient clipping at norm 1.0, and one epoch with 32 items per step (750 steps for 24,000 items), in bf16 with gradient checkpointing. Items whose prompt exceeds 1,024 tokens are dropped, not truncated (4 to 26 per run, and at most 112 in the task-transfer sets). The loss is the cross-entropy over the option keys under the read-out described above. Runs with rank 64 or 256 use $\alpha = 2r$. Full-parameter runs train all 26.9B language-model weights with the same schedule, batch size, and clipping, across eight GPUs with FSDP, fp32 master weights, and bf16 compute. The seed is 0, and 1 and 2 for the replicates of the three-seed settings. The training set of a setting is always drawn with seed 0, so replicates differ in item order, adapter initialization, and dropout. Each adapter is merged into the base and evaluated with the engine of the untuned bases. Software: torch 2.10.0 (CUDA 12.8), transformers 5.17.0, and peft 0.21.0. Of the 81 runs, 43 were trained and evaluated on H100 80GB GPUs and 38 on B200 GPUs; Table~\ref{tab:training-checks} shows that the two agree within seed noise.

\paragraph{Training data.}
Table~\ref{tab:impl-pool} lists the pool. A training set splits a language's share equally over the sub-categories that the language has in the pool, and a sub-category's share equally over its sources; a source that runs short passes the remainder on. The English-only set holds 24,000 English items. The 13-language set holds 12,000 English items and 923 for each other language, and the 11-language set 12,000 English items and 1,091 for each language except Tamil and Swahili. A dose set replaces $K$ English items of the English-only set with $K$ items of the target language. The task-transfer sets add English items from GSM8K, MS~MARCO, When2Call, and Davidson for the four unseen sub-categories, and the 96k and 288k sets scale the 13-language recipe up.

\begin{table}[H]
\centering\footnotesize
\begin{tabular}{@{}llrl@{}}
\toprule
\textbf{Sub-cat.} & \textbf{Source (training split)} & \textbf{Items} & \textbf{Languages} \\
\midrule
1.1 & ARC & 3,361 & en \\
1.1 & Global-MMLU (dev) & 3,404 & en, zh, es, ar, hi, ru, de, fr, sw, id, vi, ms \\
2.1 & COPA & 400 & en \\
2.1 & XCOPA (validation) & 500 & zh, sw, id, vi, ta \\
2.2 & XStoryCloze & 2,880 & en, zh, es, ar, hi, ru, sw, id \\
3.1 & IndicXNLI & 20,000 & ta \\
3.1 & MNLI & 20,000 & en \\
3.1 & XNLI & 220,000 & en, zh, es, ar, hi, ru, de, fr, sw, vi, th \\
3.2 & PAWS-X & 100,000 & en, zh, es, de, fr \\
3.3 & TweetSentiment-multilingual & 11,030 & en, es, ar, hi, de, fr \\
3.3 & Wisesight & 20,000 & th \\
4.1 & RACE & 20,000 & en \\
4.2 & SIB-200 & 9,814 & all 14 \\
5.1 & MASSIVE & 155,230 & all 14 \\
7.1 & PolyGuardMix & 180,000 & en, zh, es, ar, hi, ru, de, fr, th \\
\midrule
 & Total & 766,619 & \\
\bottomrule
\end{tabular}
\caption{The training pool of the controlled runs: public training, development or validation splits in the evaluation format, at most 20,000 items per source and language, after removing 452 items that overlap the evaluation set. Sub-category numbers as in Table~\ref{tab:benchmarks}.}
\label{tab:impl-pool}
\end{table}

\paragraph{Statistics.}
All resampling uses seed 0. Confidence intervals come from 1,000 bootstrap replicates paired over items. Temperature scaling fits one scalar $T$ per model and language, with $p_T \propto p^{1/T}$, by minimizing the negative log-likelihood of the gold option over $T \in [0.05, 20]$ with a golden-section search on $\log T$. The fitting rows are drawn at random 50 times and never scored, and 500 rows per language are reserved so that every fitting size is scored on the same rows. The $p$-value of the variance decomposition comes from 200,000 simulated draws.

\end{document}